\documentclass[11pt]{article}
\usepackage{url}
\usepackage{times}
\usepackage{latexsym}

\usepackage[T1]{fontenc}
\usepackage[utf8]{inputenc}

\usepackage{microtype}

\usepackage{inconsolata}

\usepackage{graphicx}
\usepackage{xcolor}

\usepackage{hyperref}
\definecolor{cornflowerblue}{rgb}{0.39, 0.58, 0.93}
\hypersetup{
    colorlinks=true,
    linkcolor=cornflowerblue,
    filecolor=magenta,      
    urlcolor=teal,
    citecolor=cornflowerblue,%teal,
    pdfpagemode=FullScreen,
    }

\usepackage{subcaption}

\usepackage{xcolor}
\usepackage{makecell}
\usepackage{enumitem}
\usepackage{xspace}
\usepackage{adjustbox}

\usepackage{iclr2025_conference}
\newsavebox{\largestimage}

\usepackage{listings}

\usepackage{amsmath}
\usepackage{mathtools}
\usepackage{multirow}

\usepackage[capitalize,noabbrev]{cleveref}

\title{M2R2: Mixture of Multi-Rate Residuals for Efficient Transformer Inference}
\author{Nikhil Bhendawade\thanks{Correspondence to: Nikhil Bhendawade <nbhendawade@apple.com>}, Mahyar Najibi, Devang Naik, Irina Belousova \\
Apple\\
\texttt{\{nbhendawade, najibi, naik.d, ibelousova\}@apple.com}
}
\iclrfinalcopy
\begin{document}

\maketitle

\begin{abstract}
Residual transformation is critical to improving representational depth and expressive power of large language models (LLMs). However, the use of static residual transformations across all tokens during auto-regressive generation induces a sub-optimal balance between inference efficiency and generation fidelity. Existing methods, including Early Exiting, Skip Decoding, and Mixture-of-Depth, attempt to address this by modulating the residual transformation based on token-level complexity. Nevertheless, these approaches predominantly consider the distance traversed by tokens through the model layers, neglecting the underlying velocity of residual evolution. In this work, we introduce \textit{Mixture of Multi-rate Residuals}, a novel framework that dynamically modulates the velocity of residual transformations to optimize early residual alignment. This modification improves inference efficiency by better aligning intermediate representations at earlier stages.

We show the efficacy of our technique in diverse optimization setups such as dynamic computing, speculative decoding, and MoE Ahead-of-Time (AoT) loading using challenging reasoning tasks from Koala, Self-Instruct, WizardLM and MT Bench. Our approach empirically outperforms state-of-the-art distance-based residual strategies, enabling a better trade-off between generation metrics and speedup in dynamic computing settings. In self-speculative decoding setups, M2R2 achieves up to 2.8X speedups on MT-Bench under lossless conditions, outperforming SOTA approaches such as 2-model speculative decoding, Medusa, LookAhead Decoding, and DEED. In Mixture-of-Experts (MoE) architectures, we enhance decoding speed by coupling early residual alignment with ahead-of-time expert loading into high-bandwidth memory (HBM). This enables concurrent memory access and computation, reducing the latency bottlenecks inherent in expert switching during decoding. Empirical results show that our method delivers a speedup of 2.9X in MoE architectures, positioning it as a highly effective strategy in resource-constrained environments.

% We validate these improvements across a broad spectrum of supervised fine-tuning and instruction-tuning tasks, achieving a superior trade-off between generation quality and inference efficiency relative to SOTA dynamic compute approaches.
 
\end{abstract}

\section{Introduction}
\label{sec:intro}
Large Language Models (LLMs) have become a cornerstone of contemporary natural language processing (NLP) systems, exhibiting exceptional performance in tasks requiring the understanding and generation of complex, structured language across diverse domains \cite{brown2020language, radford2019language, vaswani2017attention}. Their success largely stems from their ability to capture long-range dependencies in text, enabling the modeling of intricate linguistic patterns and semantics. A critical architectural component enabling this capability is the residual transformation mechanism, which introduces a pathway for the direct flow of information across layers. By preserving representations from earlier layers and allowing new transformations to build on them, residual connections help mitigate the degradation of critical information, thereby expanding the model’s expressive power to handle more abstract and nuanced features \cite{he2016identity, ba2016layer, vaswani2017attention}. This architecture empowers LLMs to model complex dependencies over extended sequences, thereby improving feature extraction and representational depth. However, the use of a static residual transformation for all tokens creates a rigid balance between inference efficiency and generation quality \cite{shen2021dynamic, garncarek2021dynamic}. This approach fails to account for the inherent variability in token complexity, leading to inefficiencies in dynamic compute scenarios.

\begin{figure}[ht]
    \centering
    \begin{subfigure}{0.49\textwidth}
        \centering
        \includegraphics[width=\textwidth]{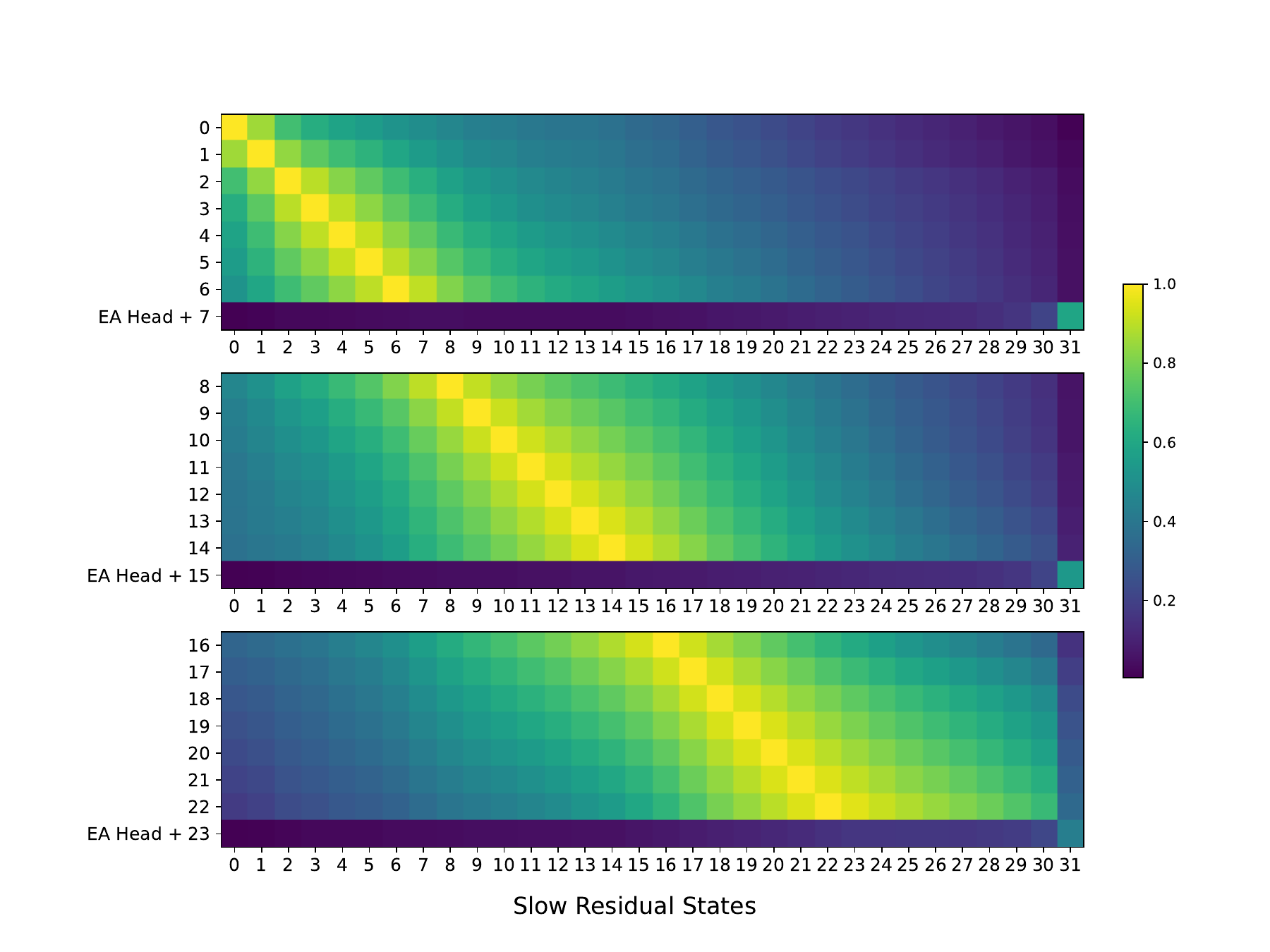}
        \caption{Residual similarity in traditional early exiting}
        \label{fig:vanilla_ea_residual_sim}
    \end{subfigure}%
    \hfill
    \begin{subfigure}{0.49\textwidth}
        \centering
        \includegraphics[width=\textwidth]{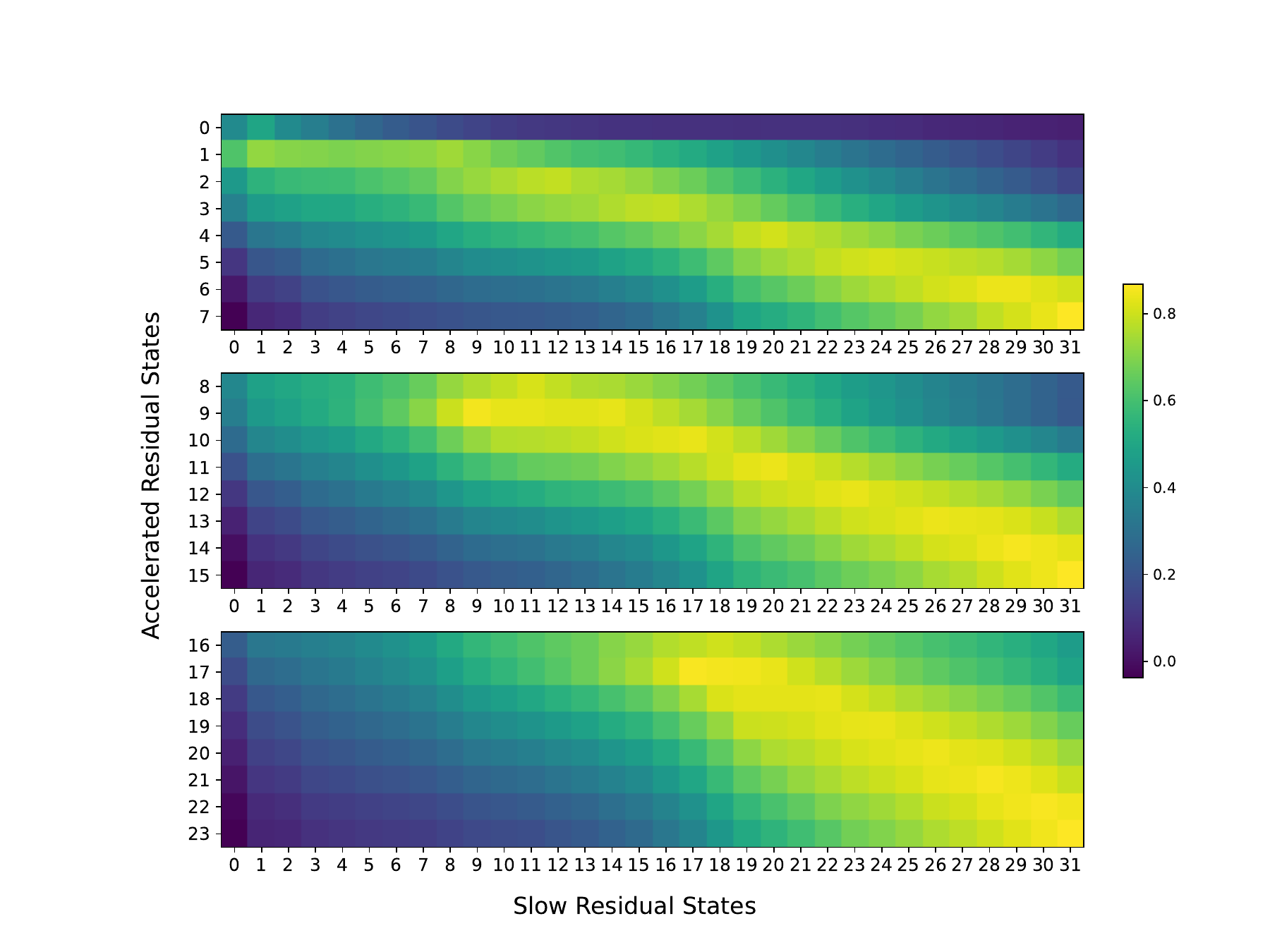}
        \caption{Residual similarity with M2R2}
        \label{fig:m2r2_residual_sim}
    \end{subfigure}
    \caption{Traditional early exiting approaches approximate the final residual state with context-independent mapping, $\mathcal{T}$, applied on intermediate hidden state, resulting in discontinuities in transformations and lower similarity with final residual state. In contrast, M2R2 progressively enhances residual transformation velocity at each layer, enabling more robust and uniform early alignment.}
    \label{fig:residual_sim}
\end{figure}

Recent initiatives to rectify this limitation have centered on the introduction of adaptive computation mechanisms. Approaches such as Early Exiting \cite{schuster2022confident, varshney-etal-2024-investigating, chen2023eellm}, Skip Decoding \cite{delcorro2023skipdecode}, Mixture-of-Depth \cite{raposo2024mixture} aim to mitigate computational overhead by dynamically modulating the depth of residual transformations based on the complexity of individual tokens. While these methodologies have demonstrated promise in enhancing inference efficiency, they predominantly focus on the distance that a token traverses through the model layers—specifically, the number of layers a token traverses prior to exiting. This distance-centric perspective neglects a critical dimension of residual transformations: the velocity at which token representations evolve within the model. In this context, velocity denotes the rate at which the residuals transform as tokens navigate the network, a factor we hypothesize can be leveraged to achieve an optimal balance between generation quality and computational efficiency.

In this paper, we introduce a novel framework, \textit{Mixture of Multi-rate Residuals}, which emphasizes the modulation of residual transformation velocity to improve early residual alignment. This enhancement boosts efficiency across diverse inference paradigms, including dynamic computing, speculative decoding, and Mixture-of-Experts (MoE) Ahead-of-Time (AoT) loading. In dynamic compute settings, rather than solely adjusting the computational depth for tokens ~\cite{chen2023eellm, schuster2022confident, delcorro2023skipdecode, Tang2024}, our method explicitly accelerates the rate of residual transformation, allowing for faster alignment of token representations at earlier stages of computation. In self-speculative decoding ~\cite{leviathan2023fast, chen2023accelerating}, accelerated residual streams facilitate the generation of speculative tokens, which are validated in parallel against slower residual streams. This parallelism enables the the advancement of multiple tokens per forward pass at the expense of loading all model parameters. Furthermore, our method also enables efficient inference of Mixture-of-Experts (MoE) models ~\cite{shazeer2017outrageously, fedus2022switch} in resource constrained setups. While sparse MoE architectures decrease the number of active parameters per decoding step, the experts necessary for a token can only be identified immediately before computation via routers that utilize the residual state to output the required expert probabilities. This necessitates expert loading from low bandwidth memory (LBM) to the accelerator, which becomes a significant bottleneck ~\cite{lepikhin2020gshard, fedus2022switch}. In contrast, our method allows for the early speculation of experts through accelerated residuals, enabling temporal overlap between computation and memory transfer, thereby reducing latency associated with expert switching during inference. 

% Our empirical results demonstrate a \textbf{<>}\% improvement in decoding speed, rendering our approach highly effective in resource-constrained environments.

To substantiate our claims, we conduct comprehensive evaluations across a spectrum of reasoning oriented and application specific tasks. Our findings illustrate that the Mixture of Multi-rate Residuals framework provides practical benefits in various inference optimization scenarios, including dynamic computation, speculative decoding, and the inference of Mixture-of-Experts (MoE) models. The contributions of this paper are as follows:
\begin{itemize}
    \item We introduce \textit{Mixture of Multi-rate Residuals} (M2R2), a novel framework that dynamically modulates residual velocity to optimize early alignment, enhancing inference efficiency and achieving a superior trade-off between decoding speedup and generation quality across diverse reasoning-oriented tasks—all without requiring costly pre-training.
    \item We empirically demonstrate that M2R2 surpasses state-of-the-art self-speculative decoding methods, attaining a 2.8$\times$ speedup on MT Bench.
    \item We further establish the effectiveness of M2R2 for on-device Mixture-of-Experts (MoE) models with Ahead-of-Time (AoT) Expert Loading, which mitigates the latency associated with on-demand expert retrieval by overlapping memory transfers with computation, yielding a 2.9$\times$ speedup over traditional expert loading methods~\citep{lepikhin2020gshard, fedus2022switch}.
\end{itemize}

%\input{sections/related}
% \begin{figure}
%     \centering
%     \includegraphics[width=0.5\linewidth]{Move_teaser.pdf}
%     \caption{Comparison of different dynamic compute approaches. length of arrow indicates residual transformation per token while width indicates velocity of transformation.}
%     \label{fig:enter-label}
% \end{figure}

\section{Method}
\label{sec:method}
Residual connections play a crucial role in shaping token representations, yet their dynamics remain underexplored in the context of efficient decoding. In this work, we delve deeper into transformer residual dynamics and investigate how modulating residual transformation velocity can improve inference efficiency in token-level processing, optimizing both dense and sparse MoE transformers.

\subsection{Residual Dynamics and Motivation for Multi-rate Residuals} \label{sec:motivation}

To analyze how hidden representations evolve across different layers of a transformer architecture, it's crucial to consider the effect of residual connections. Each transformer decoder layer typically has residual connections across attention and MLP submodules. As the residual stream $h_i$ traverses from interval $E_j$ to $E_{j+1}$, it undergoes a residual transformation given by:  
% \begin{equation}
% \label{eq:slow_residual_transformation}
% H_{E_{j+1}} = H_{E_j} \prod_{i=E_j}^{E_{j+1}} \left( I + \mathcal{A}_i \right) \left( I + \mathcal{M}_i \right) \quad \text{where} \quad \mathcal{A}_i = f(c_i, h_{i}), \mathcal{M}_i = g(h_i)
% \end{equation}

\begin{equation} \label{eq:slow_residual_transformation}
h_{E_{j+1}} = h_{E_j} + \sum_{i=E_j}^{E_{j+1}-1} \left( \mathcal{A}_i(h_i) + \mathcal{M}_i(h_i + \mathcal{A}_i(h_i)) \right) \quad \text{where} \quad \mathcal{A}_i = f(c_i, h_{i}), \mathcal{M}_i = g(h_i). 
\end{equation}

Here, \( \mathcal{A}_i \) denotes the non-linear transformation introduced by the multi-head attention mechanism at layer \( i \), while \( \mathcal{M}_i \) corresponds to the non-linear transformation of the MLP block at the same layer. These transformations depend on the input residual stream \( h_i \) and, in the case of \( \mathcal{A}_i \), the previous contextual representation \( c_i \).\footnote{Normalization layers are typically applied in practice but are omitted here for simplicity of the argument.}

% For easy tokens, the magnitude and direction of this delta transformation become progressively smaller with each successive layer as shown in \cref{fig:delta_transformation}. Consequently, it is feasible to predict these tokens after only a few residual connections, whereas harder tokens necessitate more extensive processing through additional layers.

\begin{figure}[ht]
    \centering
    \begin{subfigure}{0.48\textwidth}
        \centering
        \includegraphics[width=\textwidth]{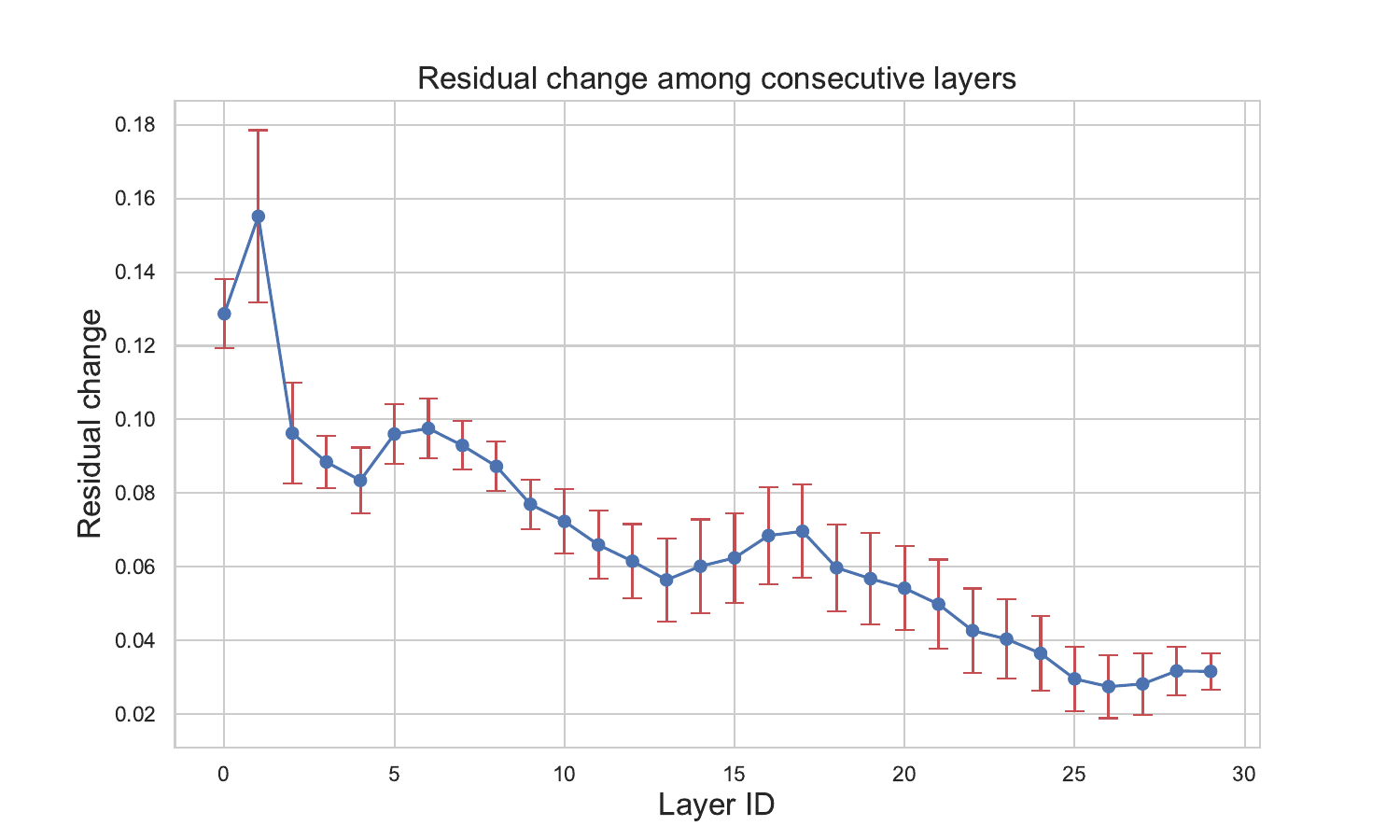}
        \caption{}
        \label{fig:residual_change}
    \end{subfigure}%
    \hfill
    \begin{subfigure}{0.48\textwidth}
        \centering
        \includegraphics[width=\textwidth]{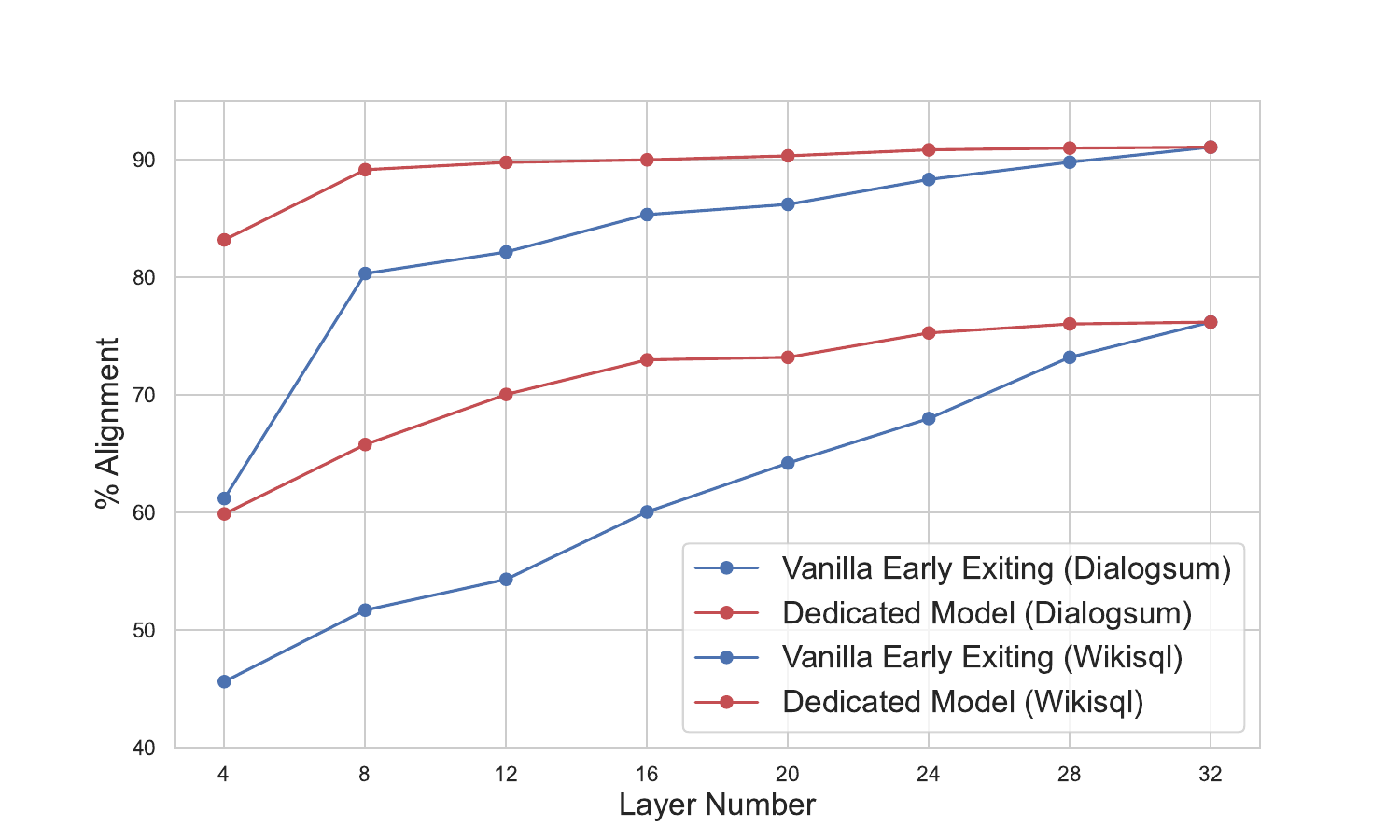}
        \caption{}
    \label{fig:alignment_wrt_dedicated_model}
    \end{subfigure}
    \caption{(a) As residual streams propagate through the model, the directional shifts in the residuals become progressively smaller. (b) A dedicated model with $k$ layers achieves a faster rate of change in residual streams and higher alignment than base model leveraging early exit mechanisms at layer $k$.}
    \label{fig}
\end{figure}

To examine whether residual transformations can be accelerated across layers, we conducted experiments using a diverse set of prompts on a pre-trained Phi3 model~\cite{phi3_report}. As illustrated in \cref{fig:residual_change}, we measured the directional shift in residual states as \( 1 - \mathcal{C}(h_{i-1}, h_i) \), where \(\mathcal{C}\) denotes normalized cosine similarity. This shift is notably higher in the initial layers, gradually decreasing in subsequent layers. This behavior allows traditional early exit approaches to effectively accelerate decoding by enabling earlier exits for simpler tokens. However, these approaches typically rely on a distance-based approximation, where the full residual transformation of the model is approximated by the residual transformations of the initial layers. To gain deeper insights into the distance versus velocity aspects of residual transformation, we conducted a comparative study. Specifically, we trained an early exit head at layer $k$ of the Phi3 model, which consists of 32 layers, restricting the distance traveled by each token. To accelerate the residual transformation relative to number of layers, we trained a smaller model consisting of only $k$ layers, while keeping all other hyperparameters consistent. We then compared the next-token prediction accuracy of the early exit head of the base model with that of the smaller model. To ensure an equal number of trainable parameters, we inserted low-rank adapters into the smaller model and trained only these adapters, whereas, in the distance-based approach, we trained solely the early exit head. In addition, to accelerate the residual transformation in smaller model, we distilled the residual streams from the larger model by incorporating a distillation loss ~\cite{sanh2019distilbert} between the residual state at layer \(i\) of the smaller model and the residual state at layer \(4 \times i\) of the larger model. As shown in ~\cref{fig:alignment_wrt_dedicated_model} the smaller model demonstrates a significantly faster rate of change in residual streams, leading to higher next token prediction accuracy after $k$ layers compared to the base model that employs traditional early exit mechanisms after $k$ layers \cite{schuster2022confident, chen2023eellm, varshney-etal-2024-investigating}. This experimental setup, which modifies only the rate of change in residual streams while keeping other factors constant, suggests that dense transformers, trained with a fixed number of layers, may inherently possess a slow residual transformation bias.

This observation raises an intriguing question: if the rate of change in residual streams could be accelerated relative to the number of layers, is it possible to facilitate earlier alignment for a greater proportion of tokens? Earlier alignment would be beneficial to not only facilitate dynamic computation but also for generating speculative tokens efficiently with high acceptance rates in speculative decoding setups ~\cite{leviathan2023fast, chen2023accelerating}. 

%thereby enhancing the efficiency of early exiting? 
 % This bias likely constrains the effectiveness of early exiting, particularly for easier tokens. By addressing this limitation through accelerated residual transformations, we hypothesize that it is possible to substantially improve the efficiency and accuracy of early exit strategies in transformer models.

\subsection{Multi-Rate Residual Transformation} \label{m2r2_method}

To address the slow residual transformation bias described in ~\cref{sec:motivation}, we introduce \textit{accelerated residual streams} that operate at rate $R$ relative to original slow residual stream. We pair slow residual stream, $h$ with an accelerated residual stream, $p$, which has an intrinsic bias towards earlier alignment. Relative to ~\cref{eq:slow_residual_transformation}, accelerated residual transformation from interval $E_j$ to $E_{j+1}$ can be represented as: 

% \begin{equation}
% \label{eq:fast_residual_transformation}
% P_{E_{j+1}} = P_{E_j} \prod_{i=E_j}^{E_{j+1}} \left( I + \hat{\mathcal{A}_i} \right) \left( I + \hat{\mathcal{M}_i} \right) \quad \text{where} \quad \hat{\mathcal{A}_i} = \hat{f}(c_i, P_{i}), \hat{\mathcal{M}_i} = \hat{g}(P_{i})
% \end{equation}

\begin{equation} \label{eq:fast_residual_transformation}
p_{E_{j+1}} = p_{E_j} + \sum_{i=E_j}^{E_{j+1}-1} \left( \hat{\mathcal{A}_i}(p_i) + \hat{\mathcal{M}_i}(p_i + \hat{\mathcal{A}_i}(p_i)) \right) \quad \text{where} \quad \hat{\mathcal{A}_i} = \hat{f}(c_i, p_{i}), \hat{\mathcal{M}_i} = \hat{g}(h_i), 
\end{equation}

where $\hat{\mathcal{A}_i}$ and $\hat{\mathcal{M}_i}$ denote non-linear transformation added by layer $i$ to previous accelerated residual $p_{i}$. Similar to $\mathcal{A}_i$, non-linear transformation $\hat{\mathcal{A}_i}$ attends to same context $c_i$ but uses a different transformation $\hat{f}$ for accelerating $p_{E_j}$ relative to $h_{E_j}$. 

\begin{figure}
    \centering
    \includegraphics[width=0.8\linewidth]{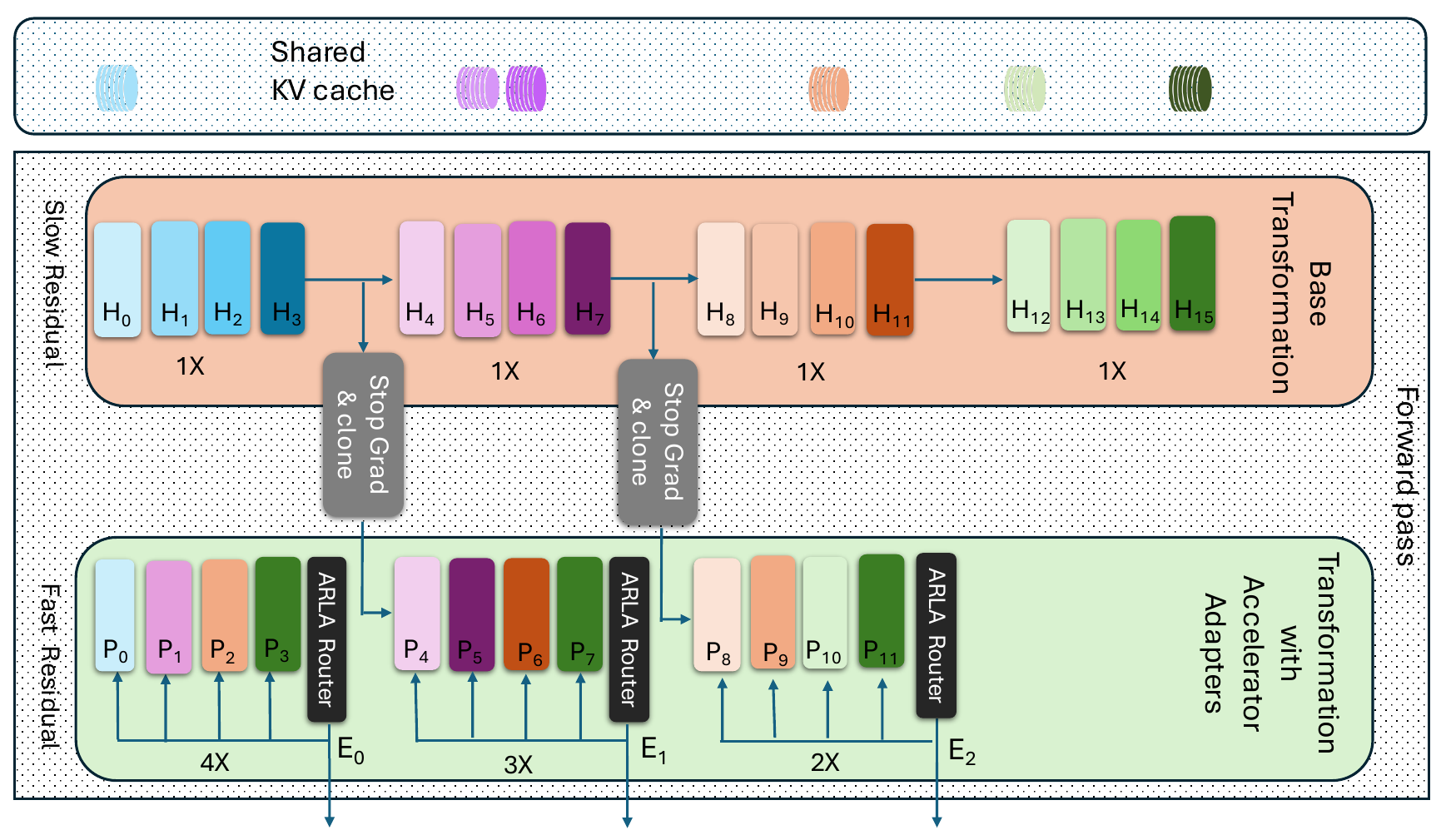}
    \caption{Multi-rate Residuals Framework: Slow residual stream of base model is accompanied by a faster stream that operates at a $2-(J+1)\times$ rate relative to the slow stream, undergoing transformations via accelerator adapters as detailed in \cref{m2r2_method}, where J denotes number of early exit intervals. Colors within the slow and fast residual streams indicate similarity, with matching colors representing the most closely aligned residual states. At the beginning of the forward pass and at each exit point, the accelerated residual state is initialized from the corresponding slow residual state to avoid gradient conflict during training (see ~\cref{sec:grad_conflict}). Early exiting decisions are informed by the Accelerated Residual Latent Attention (ARLA) mechanism, described in \cref{method_arla}, which evaluates residual dynamics across consecutive exit gates.}
    \label{fig:m2r2_main}
\end{figure}

We integrate accelerated residual transformation directly into the base network using parallel accelerator adapters such that rank of accelerator adapters $R_p << d$ where $d$ denotes base model hidden dimension. This setup allows the slow residual stream $h_{E_j}$ to pass through the base model layers while the accelerated residual stream $p_{E_j}$ utilizes these parallel adapters as shown in ~\cref{fig:m2r2_main}. Both slow and accelerated residuals are processed in same forward pass via attention masking and incur negligible additional inference latency in memory bound decoding setups, while in compute bound decoding setups where FLOPs optimization is essential, accelerated residual stream utilizes a fraction of attention heads that of slow residual (see ~\cref{sec:flops_optimization}). Additionally, to maximize the utility of accelerated residual transformations without introducing dedicated KV caches, we propose a shared caching mechanism between the slow and accelerated streams which minimally impact alignment benefits of our approach while offering substantial memory savings (see ~\cref{fig:koala_alignment}). Specifically, the attention operation on the slow residuals \( \text{MHA}(h_t, h_{\leq t}, h_{\leq t}) \) is redefined for accelerated residuals as 
\[
\hat{\mathcal{A}} = MHA(p_t, h_{<t} \oplus p_t, h_{<t} \oplus p_t),
\]
where the accelerated residual at time-step $t$, \( p_t \) attends to the slow residual’s KV cache, facilitating the reuse of contextual information across both residual streams without incurring additional caching costs. Here, \(MHA(q, k, v) \) represents multi-head attention between query \( q \), key \( k \), and value \( v \).

% Furthermore. to maximize the benefits of fast residual transformations without using dedicated KV caches, we propose sharing the fast network’s cache with the slow network. Formally speaking, We modify attention operation on slow residuals $MHA(H_t, H_{<=t}, H_{<=t})$ as $MHA(P_{t}, H_{<t} \oplus P_t, H_{<t}  \oplus P_t)$ such that accelerated residuals attend to previous slow context KV cache, where $MHA(q,k,v)$ denotes multi head attention between query, $q$, key $k$ and value $v$.

\subsection{Enhanced Early Residual Alignment}
Early residual alignment is instrumental in optimizing early exiting, speculative decoding, and Mixture-of-Experts (MoE) inference mechanisms. In this section, we provide a detailed analysis of how accelerated residuals enhance these inference setups.

% By aligning the residual states of intermediate layers with the final output representations, the model can maintain high prediction accuracy even when computations are truncated at earlier layers. This enables more reliable early exiting, reducing the overall computational cost while preserving performance. Additionally, in speculative decoding, early residual alignment allows the model to make confident predictions using faster, partial computations, thereby accelerating inference without sacrificing output quality.

\subsubsection{Early Exiting} \label{method_early_exiting}

A prevalent strategy for enabling early exiting at an intermediate layer $E_{j}$ involves approximating the residual transformation between $E_{j}$ and the final layer $N-1$ using a linear, context independent mapping, $\mathcal{T}$, such that $H_{N-1} \approx \mathcal{T}(H_{E_{j}})$. This approximation has been extensively employed in conventional approaches ~\cite{schuster2022confident, chen2023eellm, varshney-etal-2024-investigating}, providing a computationally efficient means to project the output of deeper layers from intermediate states. Specifically, residual state of layer $N-1$ with this approximation can be expressed as:

% \begin{equation}
% \label{eq: vanila_ea_assumption}
% \Phi(H_{E_{j}}) \sim H_{E_{j}} \prod_{i=E_{j}}^{N}\left( I + \mathcal{A}_i \right) \left( I + \mathcal{M}_i \right) \quad \text{where} \quad \Phi \perp C
% \end{equation}

\begin{equation} \label{eq:early_exiting}
h_{E_j} + \sum_{i=E_j}^{N-1} \left( \mathcal{A}_i(h_i) + \mathcal{M}_i(h_i + \mathcal{A}_i(h_i)) \right) \sim \mathcal{T}(h_{E_{j}})  \quad \text{where} \quad \mathcal{T} \perp c. 
\end{equation}

Here, $\mathcal{A}_i$ and $\mathcal{M}_i$ represent the residual contributions of the multi-head attention and MLP layers, respectively, while $\mathcal{T}$ remains independent of $c$, the preceding context.

This approach is inherently limited by two major factors: first, the assumption of linearity between $h_{E_{j}}$ and $h_{N-1}$ may not hold uniformly for all tokens, particularly when $E_j \ll N$. Second, the linear transformation $\mathcal{T}$ disregards the influence of the context $c$ and fails to account for the latent representations of previous contextual states. In contrast, M2R2 accelerated residual states mitigate both of these challenges by approximating the slow residual transformation of all layers via a faster residual transformation of fewer layers as:
% \begin{equation}
% H_{E_j} \prod_{i=E_j}^{N}\left( I + \mathcal{A}_i \right) \left( I + \mathcal{M}_i \right) \sim P_{E_j} \prod_{i=E_j}^{E_j+1}\left( I + \hat{\mathcal{A}_i} \right) \left( I + \hat{\mathcal{M}_i} \right)
% \end{equation}

\begin{equation} \label{eq:m2r2_approximating_ea}
h_{E_j} + \sum_{i=E_j}^{N-1} \left( \mathcal{A}_i(h_i) + \mathcal{M}_i(h_i + \mathcal{A}_i(h_i)) \right) \sim p_{E_j} + \sum_{i=E_j}^{E_{j+1}-1} \left( \hat{\mathcal{A}_i}(p_i) + \hat{\mathcal{M}_i}(p_i + \hat{\mathcal{A}_i}(p_i)) \right), 
\end{equation}

% \begin{equation} \label{eq:fast_residual_transformation}
% p_{E_{j+1}} = p_{E_j} + \sum_{i=E_j}^{E_{j+1}-1} \left( \hat{\mathcal{A}_i}(p_i) + \hat{\mathcal{M}_i}(p_i + \hat{\mathcal{A}_i}(p_i)) \right) \quad \text{where} \quad \hat{\mathcal{A}_i} = \hat{f}(c_i, p_{i}), \hat{\mathcal{M}_i} = \hat{g}(h_i) 
% \end{equation}

where $p_{E_j}$ is initialized from the slow residual state $h_{E_j}$ at each early exit interval $E_j$ using an identity transformation (see ~\cref{fig:m2r2_main}). As shown in ~\cref{fig:m2r2_residual_sim}, accelerated residuals offer a smoother, more consistent shift in residual direction across layers, in contrast to the abrupt changes typically seen at early exit points in standard early exit methods. Moreover, the normalized cosine similarity between accelerated states at early exit intervals and final residual states is substantially higher compared to traditional early exit techniques, highlighting improved alignment with final layer representations. Traditional adaptive compute methods are constrained by two principal factors: the number of tokens eligible for early exit at intermediate layers and the precision of early exit decision. If residual streams fail to saturate early, the majority of tokens remain ineligible for exit, thereby diminishing potential speedups. Additionally, imprecise delineations between tokens suitable for early exit can lead to underthinking (premature exits that adversely affect accuracy) or overthinking (unnecessary processing that compromises efficiency) ~\cite{zhou2020self, dai2020dynamic}. Enhanced early alignment using ~\cref{eq:m2r2_approximating_ea} helps to address  first issue. To address the second issue we introduce Accelerated Residual Latent Attention, which dynamically assesses the saturation of the residual stream, allowing for a more precise differentiation between tokens that can exit early and those requiring further processing.

% This results in uniform change in residual direction    
% % We keep $\mathcal{A} = \hat{\mathcal{A}}$, while $\hat{\mathcal{M}}$ is accelerated by a factor of $2 - (N_{E}+1)X$ relative to the slower residual transformation $\mathcal{M}$, where $N_E$ represents number of early exiting intervals.
% Figure~\cref{fig:rate_change_comparison} illustrates the comparative rate of change between these transformation streams.

% fig:rate_change_comparison
% - grid plot x axis -> layer id (0, 8) , y axis -> layer id -> dark color cell for max similarity , lighter for lower 
% \input{sections/equations.tex}

% \begin{figure}[ht]
%     \centering
%     \includegraphics[width=0.8\textwidth, height=5cm]{rate_change_comparison.png}
%     \caption{Rate of change comparison between fast and slow residual streams.}
%     \label{fig:rate_change_comparison}
% \end{figure}

%vary k and and plot EA accuracy for larger and smaller models. 

% \begin{figure}[ht]
%     \centering
%     \includegraphics[width=0.5\textwidth,height=5cm]{sections/figures/alignment_comparison_dialogsum.pdf}
%     \caption{Alignment of exited tokens for different early exit layers using traditional early exiting heads, dedicated faster networks, and faster residuals.}
%     \label{fig:small_model_early_exiting}
% \end{figure}

\textbf{Accelerated Residual Latent Attention} \label{method_arla}

\begin{figure}[ht]
    \centering
    \begin{subfigure}{0.52\textwidth}
        \centering
        \includegraphics[width=\textwidth, height = 4cm]{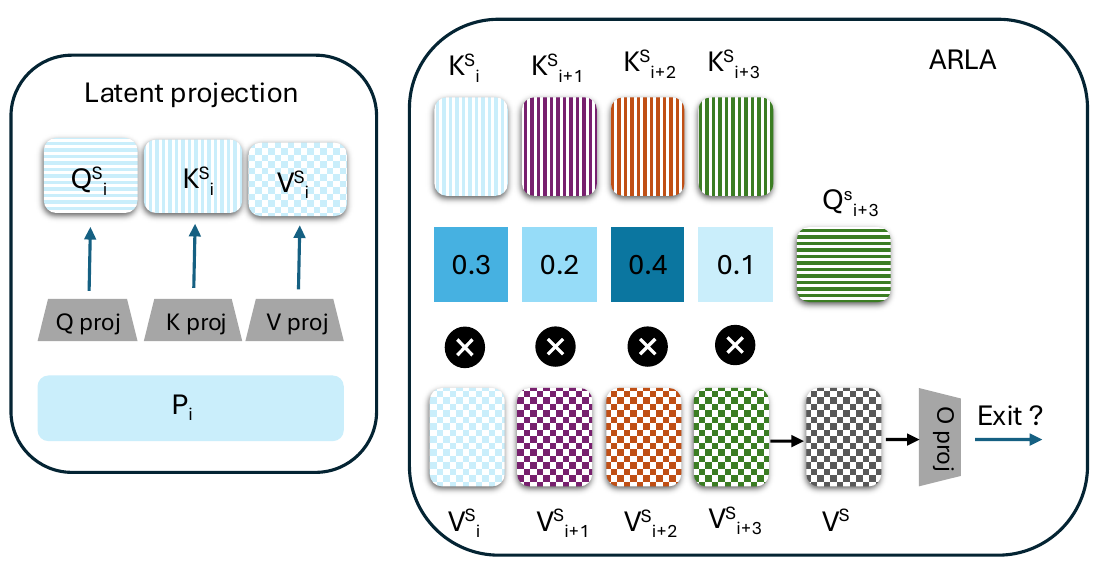}
        \caption{Accelerated Residual Latent Attention (ARLA): Accelerated residuals between early exit gates are projected into latent domain and attention over residual states within the interval is computed to capture residual dynamics and exit decision is made based on residual saturation.}
        \label{fig:arla_arch}
    \end{subfigure}%
    \hfill
    \begin{subfigure}{0.45\textwidth}
        \centering
        \includegraphics[width=\textwidth, height = 4.5cm]{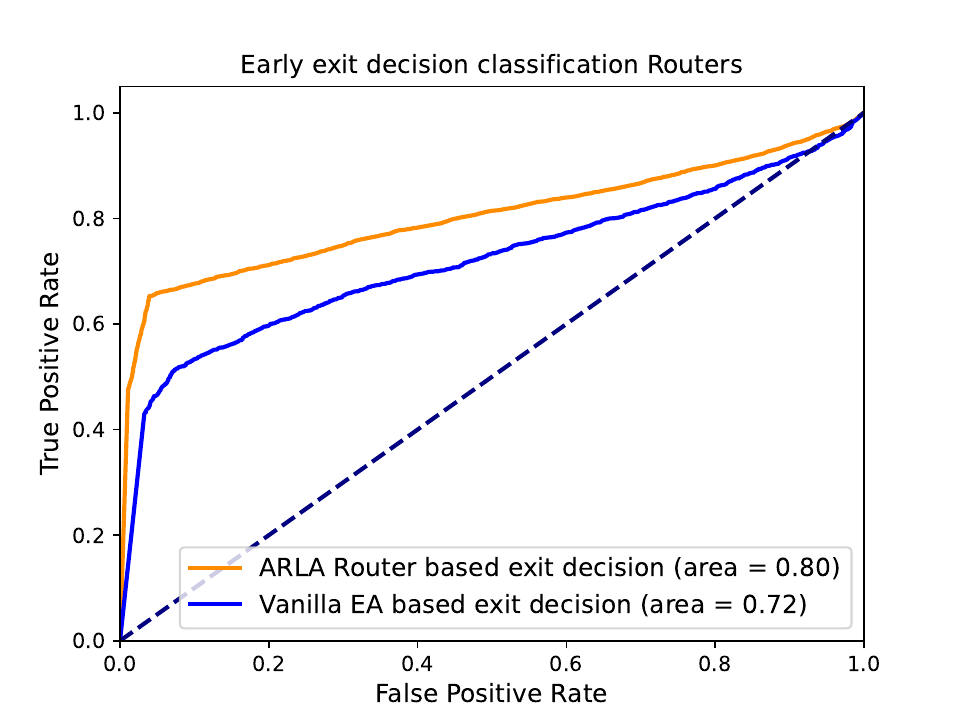}
        \caption{ROC classification curves of early exit decision strategies using a linear router used on last residual state ~\cite{schuster2022confident, varshney-etal-2024-investigating, chen2023eellm}  and using ARLA approach that considers residual dynamics. }
        \label{fig:roc_arla}
    \end{subfigure}
    \caption{Effectiveness of ARLA in capturing residual dynamics for early exiting decisions.}

\end{figure}

In the context of residual streams, we observe that the decision to exit at a given layer can be more effectively informed by analyzing the dynamics of residual stream transformations, instead of solely relying on a classification head applied at the early exit interval $E_j$. To capture the subtle dynamics of residual acceleration, we propose a \textit{Accelerated Residual Latent Attention} (ARLA) mechanism. This approach involves making the exit decision at gate $E_j$ by attending to the residuals spanning from gate $E_{j-1}$ to $E_j$, rather than considering only the residual at gate $E_j$. To minimize the computational overhead associated with exit decision-making, the attention mechanism operates within the latent domain as depicted in ~\cref{fig:arla_arch}. Formally, for each interval $[E_j, E_{j+1}]$, the accelerated residuals are projected into Query ($Q^s_{E_j}, \ldots, Q^s_{E_{j+1}}$), Key ($K^s_{E_j}, \ldots, K^s_{E_{j+1}}$), and Value ($V^s_{E_j}, \ldots, V^s_{E_{j+1}}$) vectors, with latent dimension $d^s$ for $Q^s$, $K^s$, and $V^s$ being significantly smaller than hidden dimension of $p$.\footnote{We use $d^s = 64$ for experiments described in ~\cref{sec:experiments}.} Notably, when the router is allowed to make exit decisions at gate $E_j$ based on residual change dynamics, we observe that the attention is not confined to the residual state at $E_j$ but is distributed across residual states from $E_{j-1}$ to $E_j$, %as illustrated in Figure~\ref{fig:vertical_latent_attention_dynamics}. 
This broader focus on residual dynamics significantly reduces decision ambiguity in early exits, as demonstrated in Figure~\ref{fig:roc_arla}, which contrasts routers based on the last hidden state, and the proposed ARLA router.

%show R -> S transformation. 
%show parameter and flop overhead as compared to adapter on last hidden state.

% \begin{figure}[ht]
%     \centering
%     \includegraphics[width=0.5\textwidth,height=5cm]{sections/figures/roc_arla.pdf}
%     \caption{ROC curves of early exit decision strategies: confidence-based methods (CALM/LITE), routers based on the accelerated hidden state, and latent attention routers.}
%     \label{fig:decision_making_comparison}
% \end{figure}

% \begin{figure}[ht]
%     \centering
%     \includegraphics[width=0.5\textwidth,height=5cm]{vertical_latent_attention.png}
%     \caption{Vertical latent attention mechanism for optimizing early exit decisions by considering residuals from gate \(M\) through \(M-1\).}
%     \label{fig:vertical_latent_attention}
% \end{figure}

% \begin{figure}[ht]
%     \centering
%     \includegraphics[width=1\textwidth,height=5cm]{sections/figures/arla.pdf}
%     \caption{fig that plots 32 rows 2 cols heatmap showing attention at each gate}
%     \label{fig:vertical_latent_attention_dynamics}
% \end{figure}

\subsubsection{Self Speculative Decoding} \label{method_self_speculative_decoding}

An alternative means to exploit the early alignment properties of our approach is through the use of accelerated residual states for speculative token sampling to accelerate autoregressive decoding. Speculative decoding aims to speed up memory-bound transformer inference by employing a lightweight draft model to predict candidate tokens, while verifying speculated tokens in parallel and advancing token generation by more than one token per full model invocation \cite{leviathan2023fast, chen2023accelerating, xia2023speculative, miao2023specinfer}. Despite its effectiveness in accelerating large language models (LLMs), speculative decoding introduces substantial complexity in both deployment and training. A separate draft model must be specifically trained and aligned with the target model for each application, which increases the training load and operational complexity ~\cite{chen2023accelerating}. Additionally, this approach is resource-inefficient, as it requires both the draft and target models to be simultaneously maintained in memory during inference \cite{leviathan2023fast, chen2023accelerating}. 

One strategy to address this inefficiency is to leverage the initial layers of the target model itself to generate speculative candidates, as depicted in ~\cite{Tang2024}. While this method reduces the autoregressive overhead associated with speculation, it suffers from suboptimal acceptance rates. This occurs because the linear transformation employed for translating hidden states from layer $k$ to the final layer $N$ is typically a poor approximation, as discussed in ~\cref{sec:motivation} and ~\cref{method_early_exiting}. Our approach resolves this limitation by utilizing accelerated residuals, which demonstrate higher fidelity to their slower counterparts. By utilizing accelerated residuals operating at a rate of $N/k$, where $k$ denotes the number of layers used for candidate speculation, we are able to efficiently generate speculative tokens for decoding.\footnote{We typically set $k = 4$ to balance the trade-off between autoregressive drafting overhead and acceptance rate, as discussed in~\cref{sec:experiments}.}
 This technique not only obviates the need for multiple models during inference but also improves the overall efficiency and effectiveness of speculative decoding.

\begin{figure}
    \centering    \includegraphics[width=1\linewidth]{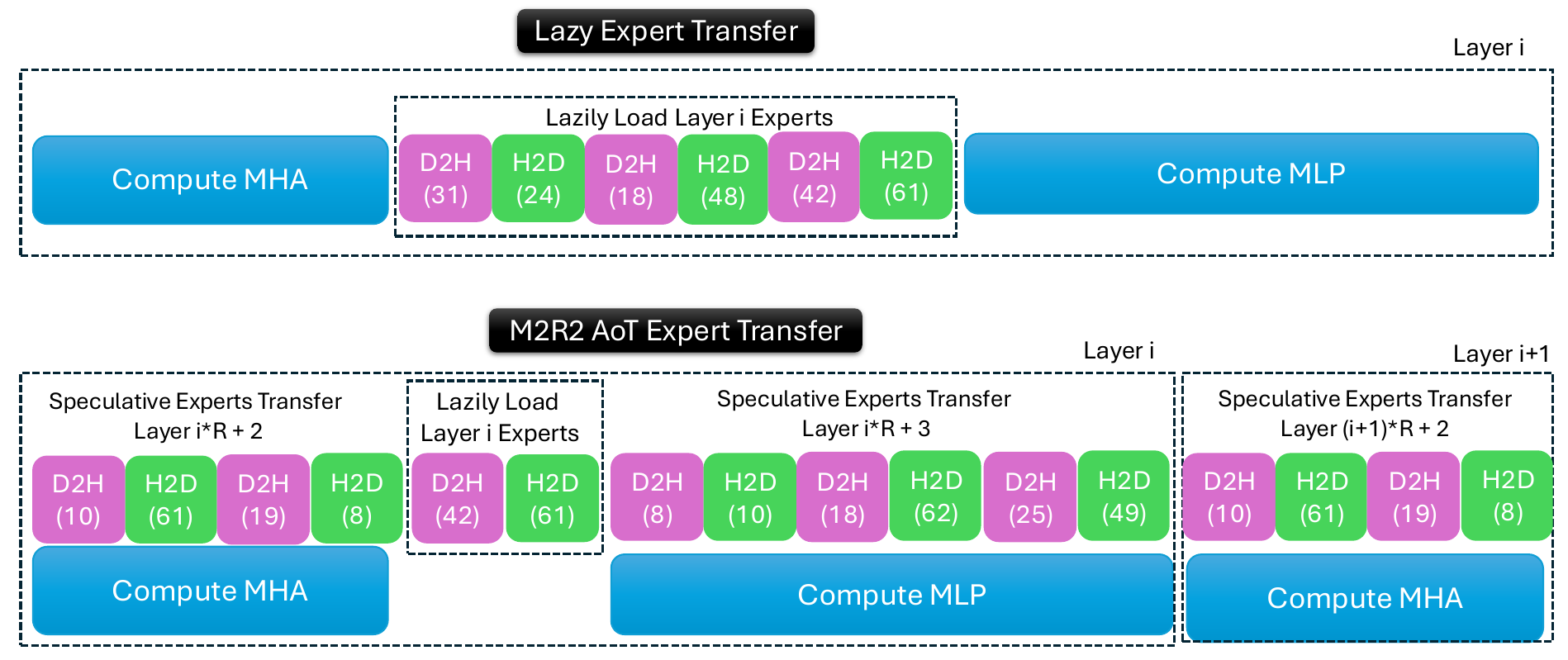}
    \caption{Ahead-of-Time Expert Loading: M2R2 accelerated residual stream predicts experts required for future layers, reducing reliance on on-demand lazy loading. Speculative pre-loading is efficiently overlapped with computation of multi-head attention (MHA) and MLP transformations. Only incorrectly speculated experts are loaded lazily, resulting in faster inference steps and improved computational efficiency. Here, H indicates LBM Host while D indicates HBM Device.}
    \label{fig:moe_expert_aot_loading}
\end{figure}

\subsubsection{Ahead of Time Expert Loading:} \label{method_aot_expert_loading}

Recent advancements in sparse Mixture-of-Experts (MoE) architectures ~\cite{shazeer2017outrageously, fedus2022switch, artetxe2019massively, lepikhin2020gshard, zoph2022designing} have introduced a paradigm shift in token generation by dynamically activating only a subset of experts per input, achieving superior efficiency in comparison to dense models, particularly under memory-bound constraints of autoregressive decoding \cite{fedus2022switch, zoph2022designing}. This sparse activation approach enables MoE-based language models to generate tokens more swiftly, leveraging the efficiency of selective expert usage and avoiding the overhead of full dense layer invocation. In dense transformer models, pre-loading layers is a common strategy to enhance throughput, as computations of current layer can be overlapped with pre-loading of next layer parameters ~\cite{narayanan2021efficient, shoeybi2020megatron}. However, MoE models face a unique challenge: expert selection occurs dynamically based on previous layer’s output, making it infeasible to preload next layer’s experts in parallel. This limitation results in inherent latency, as expert loading becomes a sequential, on-demand process ~\cite{lepikhin2020gshard, fedus2022switch}.

To address this inefficiency, our method introduces a mechanism with \textit{accelerated residuals}, which not only captures key characteristics of base slower residual states but also exhibit high cosine similarity with their final counterparts (as illustrated in \cref{fig:m2r2_residual_sim}). By employing accelerated residual streams, we can effectively predict the necessary experts for future layers well in advance of their actual invocation. Specifically, using a $2\times$ accelerated residual, the experts needed for layers $2i+2$ and $2i+3$ can be identified while still computing in layer $i$, thus overcoming the bottleneck of sequential, on-demand expert selection and mitigating latency in the decoding pipeline, as shown in \cref{fig:moe_expert_aot_loading}. Note that, we use fixed set of accelerator adapters for transforming accelerated residuals (as discussed in ~\cref{m2r2_method}) while slow residual is transformed via expert routing mechanism. 

Furthermore, our approach integrates a Least Recently Used (LRU) caching strategy, which enhances memory efficiency by replacing the least recently used experts with speculated experts that are anticipated to be needed in upcoming layers. This hybrid approach of preemptive expert loading with LRU caching yields substantial improvements over traditional on-demand loading or standalone caching strategies. By minimizing cache misses and efficiently managing memory, this approach addresses both compute and memory bottlenecks, leading to faster, more resource-efficient token generation in MoE architectures. A comprehensive evaluation of this strategy, in relation to state-of-the-art methods, is provided in \cref{experiments_aot}, and the compute and memory traces on an A100 GPU are detailed in \cref{fig:moe_aot_cuda_trace}.

\subsection{Training} \label{method_training}
% This approach is feasible due to the absence of gradient conflicts, as discussed in \cref{sec:grad_conflict}.

To accelerate residual streams, we employ parallel accelerator adapters as described in \cref{m2r2_method}.  For the early exiting use-case outlined in \cref{method_early_exiting}, we define the training objective for these adapters using the following loss function, which combines cross-entropy loss at each exit $E_j$ with distillation loss at each layer $i$. Loss weights coefficients $\alpha_0$ and $\alpha_1$ are employed to balance contribution of corresponding losses.

\begin{align} \label{eq:mr_loss}
L_{\text{m2r2}} = \underbrace{-\alpha_0 \sum_{j=1}^{J} \sum_{t=1}^{T} \log p_{\theta} \left( \hat{y}_t^{E_j} \mid y_{<t}, x \right)}_{\text{cross-entropy loss}} 
+ \underbrace{\alpha_1\sum_{i=1}^{E_{J-1}} \sum_{t=1}^{T} \| \mathbf{p}_{t}^{i} - \mathbf{h}_{t}^{((i - E_{j(i)}) \cdot R_i) + E_{j(i)})} \|^2}_{\text{distillation loss}}.
\end{align}

where $\hat{y}_t^{E_j}$ denotes the predictions from the accelerated residual stream at layer $E_j$ and time step $t$, $y_t$ represents the corresponding ground truth tokens, and $x$ indicates previous context tokens. The distillation loss at each layer $i$ is computed by comparing accelerated residuals at layer $i$ with slow residuals at layer $(i - E_{j(i)}) \cdot R_i + E_{j(i)}$, where $R_i$ denotes the rate of accelerated residuals at layer $i$ while $E_{j(i)}$ represents the most recent gate layer index such that $E_{j(i)} <= i$. \( J \) represents the total number of early exit gates, N denotes number of hidden layers and $E_j$ denotes layer index corresponding to gate index $j$ and \( T \) denotes the sequence length. 

In dynamic compute settings, after training of accelerator adapters, we optimize the query, key, and value parameters governing the ARLA routers (see ~\cref{method_arla}) across all exits in parallel on binary cross entropy loss between predicted decision and ground truth exiting decision. The ground truth labels for the router are determined based on whether the application of the final logit head on $\hat{y}_t^{E_j}$ yields the correct next-token prediction.

% The objective for this optimization is defined by the following loss function:

%TODO are equations required ? 
% \begin{equation} \label{eq:arla_loss_combined}\small
%     L_{\text{arla}} = -\frac{1}{N} \sum_{t=1}^{T} \left( \sum_{j=1}^{E_n} \left[ O_t^{E_j} \log(\hat{O}_t^{E_j}) + (1 - O_t^{E_j}) \log(1 - \hat{O}_t^{E_j}) \right] \right), \quad \text{where} \quad 
%     O_t^{E_j} = \begin{cases} 
%     1, & \text{if } L(\hat{y}_t^{E_j}) = y_t^{E_j} \\
%     0, & \text{otherwise}
%     \end{cases}
% \end{equation}

% where $\hat{O}_t^{E_j}$ represents the binary predicted logits produced by the vertical latent attention router, as described in \cref{sec:arla}, at gate $E_j$ and time step $t$, and $O_t^{E_j}$ denotes the corresponding ground truth labels. The ground truth labels for the router are determined based on whether the application of the logit head on $\hat{y}_t^{E_j}$ yields the correct next-token prediction. The parameters controlling vertical latent attention are trained concurrently to ensure consistency and efficient use of computational resources.

For self-speculative decoding, as described in \cref{method_self_speculative_decoding}, the training objective remains the same as \cref{eq:mr_loss}, but with the number of intervals set to $J = 1$ and the rate of residual transformation set to $R_n = N/k$, where the first $k$ layers generate speculative candidate tokens. In the context of Ahead-of-Time Expert Loading for Mixture-of-Experts (MoE) models (see \cref{method_aot_expert_loading}), setting the rate of residual transformation to $R_n = 2$ typically offers a good trade-off between the accuracy of expert speculation and AoT pre-loading of experts. 

% Thus, we set $J = 1$ and $E_1 = 16$.

~\subsection{FLOPs Optimization} \label{sec:flops_optimization}

Naively implemented, M2R2 incurs higher FLOP overhead compared to traditional speculative decoding and early exiting approaches such as ~\cite{medusa, schuster2022confident, Tang2024}. However, modern accelerators demonstrate compute bandwidth that exceeds memory access bandwidth by an order of magnitude or more~\cite{databricksLLMInference2023, jouppi2021ten}, meaning increased FLOPs do not necessarily translate to increased decoding latency. Nevertheless, to ensure fair comparison and efficiency in compute bound scenarios, we introduce targeted optimizations.

~\textbf{Attention FLOPs Optimization} For medium-to-long context lengths, attention computation dominates FLOPs in the self-attention layer, surpassing the contribution from MLP layers. Specifically, matrix multiplications involving queries, cached keys, and cached values scale with $l_{kv} * l_{q}$ where $l_{kv}$ denotes previous context length and $l_q$ denotes current query length. Since M2R2 pairs accelerated residuals with slow residuals, a naive implementation results in twice the FLOPs consumption compared to a standard attention layer. To address this, we limit the attention of accelerated residual stream to selectively attend to the top-k most relevant tokens, identified by the slow residual stream based on top attention coefficients\footnote{We set to k = 64 and attend to top 64 tokens as identified by the slow residual stream.}. This is possible since slow and accelerated residual streams are processed in same forward pass and accelerated streams have access to attention coefficients of slow stream. Note that, the faster residual stream still retains the flexibility to assign distinct attention coefficients to these tokens. Furthermore, we design the faster residual stream to employ only 8 attention heads, compared to the 32 heads used in the slow residual stream of the Phi-3 model, reducing query, key, value, and output projection FLOPs by a factor of 1/4. ~\cref{fig:m2r2_num_heads_ablation} indicates effect of using a slicker stream on alignment. As depicted, using $\hat{n}_h = 8$ offers a good trade-off between alignment and FLOPs overhead. 

~\textbf{MLP FLOPs Optimization} The accelerator adapters operating on the accelerated residual stream are intentionally designed with lower rank than their counterparts in the base model. This reduces FLOP overhead by a factor proportional to $hiddenSize / rank$. Additionally, since the faster residual stream uses only 8 attention heads (compared to 32 in the slow residual stream of Phi-3), the subsequent MLP layers process a smaller set of activations, further reducing FLOPs by another factor of 1/4.

These optimizations significantly reduce the FLOP overhead per speculative draft generation, as illustrated in ~\cref{fig:flops_optmization}. Notably, while traditional early-exiting speculative approaches such as DEED require propagating the full slow residual state through the initial layers, incurring substantial computational costs, M2R2 achieves efficient token generation via slimmer, low-rank faster residual streams. In contrast, Medusa introduces considerable FLOP overhead due to per-head computations scaling with $d^2+dv$\footnote{Here $d$ denotes hidden state dimension while $v$ denotes vocab size.}, whereas M2R2 employs low-rank layers for both MLP and language modeling heads, maintaining computational efficiency. All experiments involving the M2R2 approach, as detailed in ~\cref{sec:experiments}, are conducted using these FLOPs optimizations.

\section{Experiments} \label{sec:experiments}

We perform a comprehensive evaluation of our proposed method across both reasoning-intensive as well as structured application-specific tasks using pre-trained models of multiple scales. 

\textbf{Datasets.} \label{all_tasks} To assess reasoning capabilities, we adopt a comprehensive strategy by training on the instruction-tuning dataset Alpaca~\cite{touvron2023alpaca} and evaluating performance across multiple held-out human instruction test sets, including Koala~\cite{koala2023}, Self-Instruct~\cite{wang2022selfinstruct}, WizardLM~\cite{xu2023wizardlm}, and MT Bench~\cite{bai2024mtbench}. These datasets encompass a wide spectrum of instruction styles, task complexities, and domain-specific reasoning challenges, such as multi-turn interactions (Koala), open-ended problem-solving (Self-Instruct), step-by-step reasoning (WizardLM), and multi-dimensional evaluation of instruction-following capabilities (MT Bench). Beyond reasoning-oriented tasks, we further evaluate our approach on structured application-specific tasks, including Structured API Generation, Text Summarization, and Meaning Representation. For Structured API Generation, we leverage the sql-create-context dataset, which is constructed using WikiSQL~\cite{zhongSeq2SQL2017} and SPIDER~\cite{yu2018spider}; for Text Summarization, we utilize Dialogsum~\cite{chen-etal-2021-dialogsum}; and for Meaning Representation, we employ the e2e-nlg dataset~\cite{dusek_etal2020_csl}. These tasks assess the model's ability to produce well-formed outputs for practical AI assistant applications, ensuring both efficiency and applicability in real-world deployment.

\textbf{Models and Baselines.} We evaluate our proposed approach on open-source, dense transformer models of varying scales, including Phi-3-mini-4k-instruct (3.8B) \cite{phi3_report} and Gemma (7B) \cite{gemma2024}, as well as the recently introduced sparse Mixture-of-Experts (MoE) model, OlMoE (1B-7B) \cite{muennighoff2024olmoe}. To benchmark our method in dynamic compute scenarios, we compare it against several state-of-the-art dynamic compute techniques, including LITE \cite{varshney-etal-2024-investigating}, an extension of CALM \cite{schuster2022confident}, as well as skip decoding \cite{delcorro2023skipdecode} and Mixture of Depths (MoD) \cite{raposo2024mixture}. For speculative decoding scenarios, we use both standard draft-target speculative decoding method ~\cite{spector2023accelerating} and single-model baselines such as Medusa \cite{medusa}, DEED  \cite{Tang2024} and LookAhead Decoding ~\cite{fu2023lookahead}. To assess the performance of our approach in MoE configurations with Ahead-of-Time (AoT) expert loading, we conduct comparisons using several baselines. These include fixed expert configurations, LRU-based caching, LRU caching combined with random expert speculation, and expert prediction based on previous hidden states.

\textbf{Metrics} We report the trade-off between wall-time speedups and generation quality metrics on the held-out test sets to compare dynamic compute approaches. In contrast, for evaluating speculative decoding setups, we focus on wall-time speedups and acceptance rates, as speculative decoding does not affect generation quality ~\cite{spector2023accelerating}. To assess the effectiveness of our method in Ahead-of-Time (AoT) MoE expert loading, we report both expert speculation hit rate and decoding latency.

For reasoning-oriented tasks, we evaluate response quality on human instruction test sets such as Self-Instruct~\cite{wang2022selfinstruct}, Koala~\cite{geng2023koala}, WizardLM~\cite{xu2023wizardlm}, and MT Bench~\cite{zheng2023judging} using GPT-4~\cite{openai2023gpt4}. To mitigate position bias in LLM judgments~\cite{wang2022selfinstruct}, we assess response pairs in both possible orderings and aggregate the judgment scores. The prompt used for comparing response quality between models is provided in~\cref{m2r2_prompt}.  For structured application-specific tasks, we use Exact Match (EM) accuracy as the generation metric for the Structured Query task, while for Dialog Summarization, we employ the Rouge-LSum metric~\cite{wolf2020transformers}.

\textbf{Inference.} All inference experiments were conducted on a single Nvidia A100-80GB GPU, with a batch size of 1, using float16 precision and greedy decoding (temperature T = 0) to emulate common configurations for on-device AI assistants.

% For additional analyses, please refer to \cref{sec:batching} for the impact of batching, \cref{supp_ablations} for ablation studies on top-k sampling and temperature \(T = 1\), and \cref{supp_expt_details} for further experimental details.

\begin{figure}[h]
    \centering
    \begin{subfigure}[b]{0.48\textwidth}
        \centering
        \includegraphics[width=\textwidth,height=4cm]{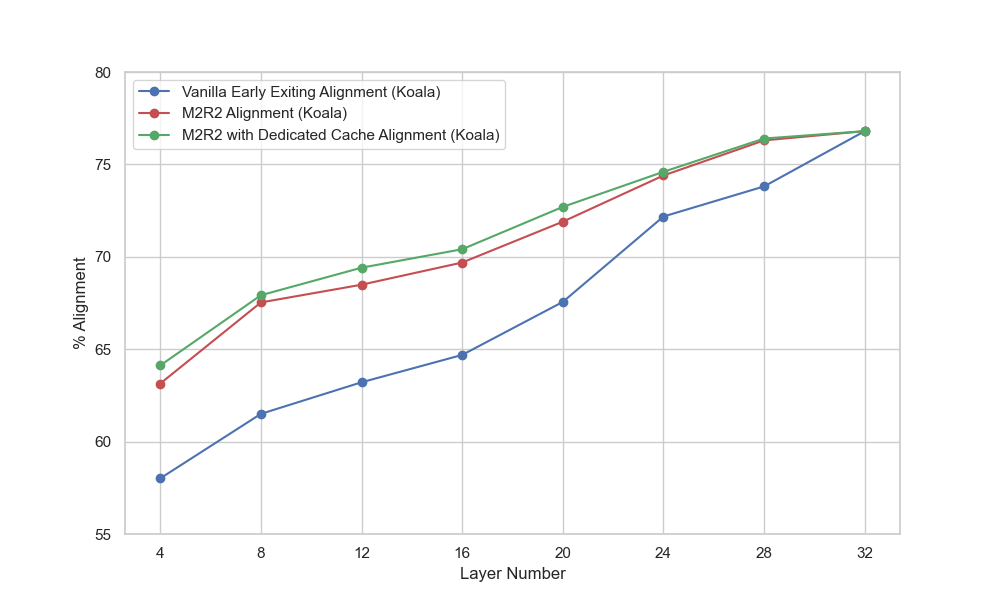}
        \caption{Early alignment performance on the Koala test set, comparing traditional early exiting with M2R2, both with and without cache sharing between slow and accelerated residuals.}
        \label{fig:koala_alignment}
    \end{subfigure}
    \hfill
    \begin{subfigure}[b]{0.48\textwidth}
        \centering
        \includegraphics[width=\textwidth,height=4cm]{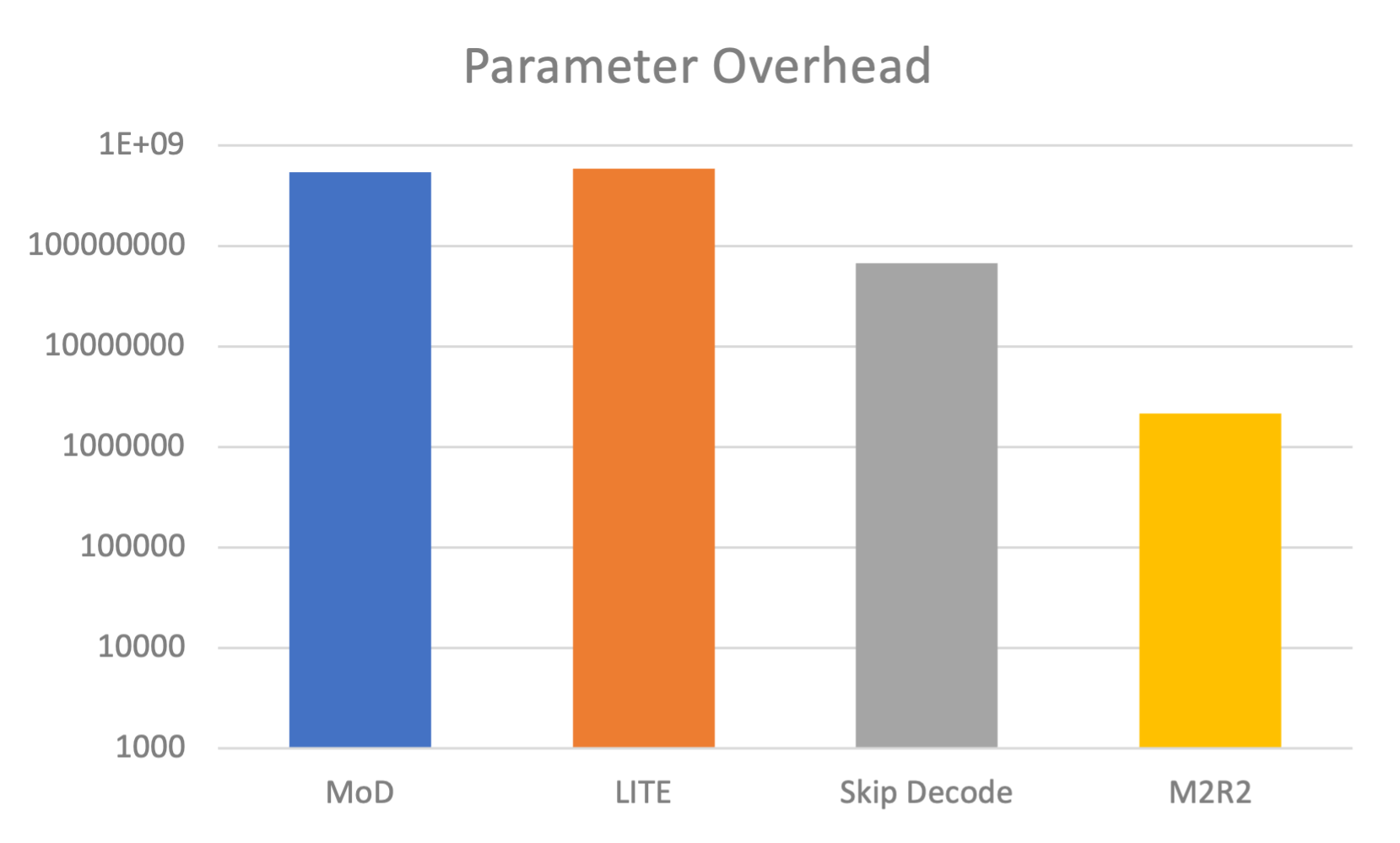}
        \caption{Parameter overhead across different dynamic compute approaches, highlighting additional trainable parameters in routers, projection layers, accelerator adapters, and ARLA.}
        \label{fig:parameter_overhead}
    \end{subfigure}
    \caption{Alignment of early exited tokens and trainable parameter overhead associated with different dynamic computing approaches.}
    \label{fig:side_by_side}
\end{figure}

\subsection{Results}

\subsubsection{Dynamic Residual Transformation}
\textbf{Early Alignment and ARLA Effectiveness} \label{early_alignment}
We begin by analyzing the alignment of tokens exited at intermediate gates with those exited at the final layer using the Koala instruction set~\cite{koala2023}. As shown in \cref{fig:koala_alignment}, we observe that accelerated residuals achieve significantly higher alignment compared to conventional early exit approaches, such as those proposed in~\cite{schuster2022confident, chen2023eellm, varshney-etal-2024-investigating}. This difference in alignment is particularly pronounced at lower gates, demonstrating that accelerated residual streams more effectively capture the features of the final-layer slow residual stream than applying a projection layer on intermediate slow residuals. Additionally, we find that sharing the KV cache between slow and accelerated residuals does not significantly impact alignment. Cache sharing allows for substantial reductions in runtime memory, and in the subsequent experiments detailed in this section, we share the cache between slow and accelerated residual streams. We also compare ROC curves obtained from confidence scores that are used to make exiting decisions in \cite{schuster2022confident} and our approach. As observed in \cref{fig:roc_arla}, ARLA described in \cref{method_arla}, is consistently effective in  optimally determining decision boundaries than classifier-based routers that operate on latest slow residual state at each gate in \cite{schuster2022confident, varshney-etal-2024-investigating}. 

\begin{figure}[ht]
    \centering
    % Row 1
    \begin{subfigure}{0.33\textwidth}
        \centering
        \includegraphics[width=\textwidth]{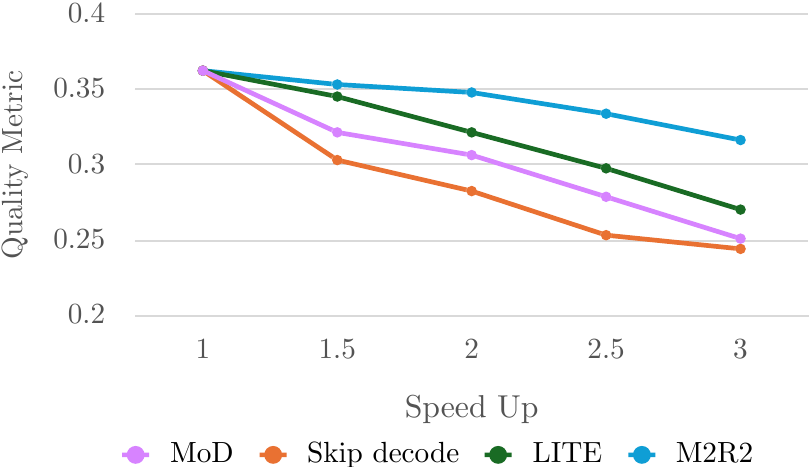}
        \caption{\small Dialog Summarization}
        \label{fig:ea_dialog}
    \end{subfigure}%
    \hfill
    \begin{subfigure}{0.33\textwidth}
        \centering
        \includegraphics[width=\textwidth]{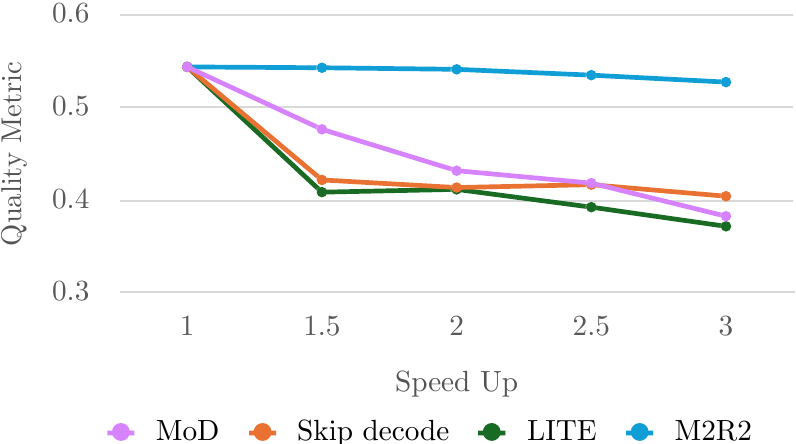}
        \caption{\small E2E-NLG}
        \label{fig:ea_meaning}
    \end{subfigure}%
    \hfill
    \begin{subfigure}{0.33\textwidth}
        \centering
        \includegraphics[width=\textwidth]{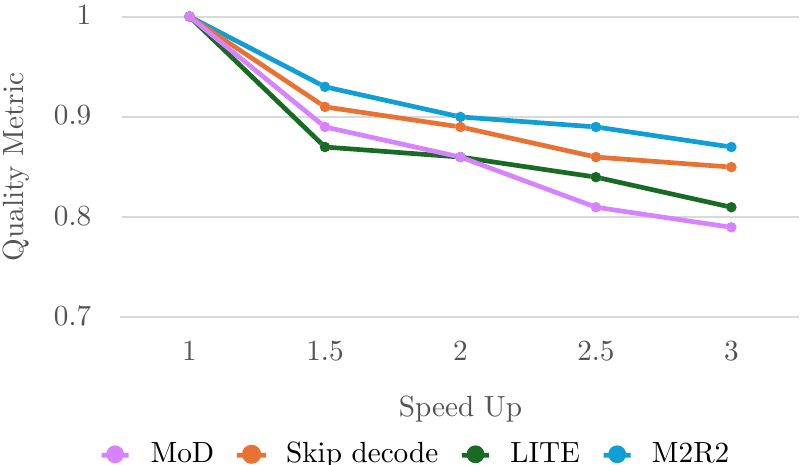}
        \caption{\small SQL Generation}
        \label{fig:ea_sql_generation}
    \end{subfigure}

    % Row 2
    \begin{subfigure}{0.3\textwidth}
        \centering
        \includegraphics[width=\textwidth]{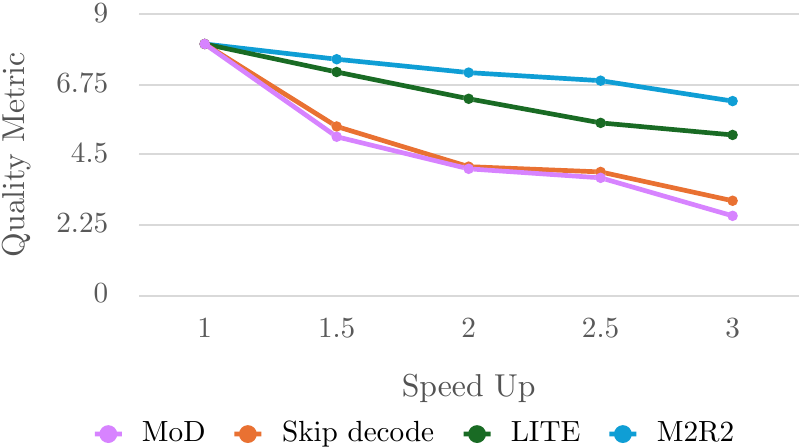}
        \caption{\small Self Instruct}
        \label{fig:ea_self_instruct}
    \end{subfigure}%
    \hfill
    \begin{subfigure}{0.3\textwidth}
        \centering
        \includegraphics[width=\textwidth]{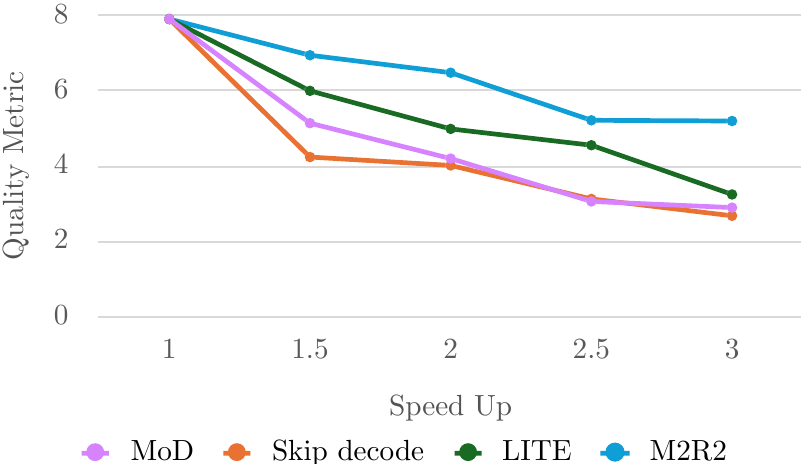}
        \caption{\small WizardLM}
        \label{fig:ea_wizardlm}
    \end{subfigure}%
    \hfill
    \begin{subfigure}{0.3\textwidth}
        \centering
        \includegraphics[width=\textwidth]{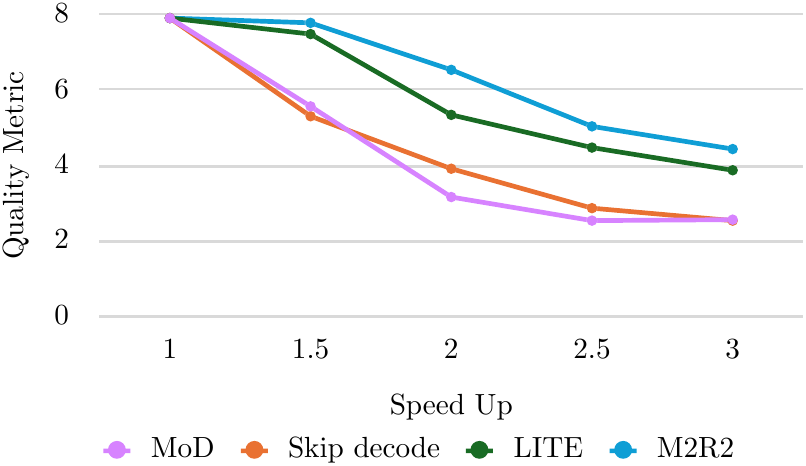}
        \caption{\small Koala}
        \label{fig:ea_koala}
    \end{subfigure}
    
    \caption{Generation Metric vs Speedup trade-off of different dynamic computing approaches with Phi-3 model on instruction and application specific test sets.}
    \label{fig:ea_comparison}
\end{figure}

 \textbf{Dynamic compute speedups}
We evaluated both LITE ~\cite{varshney-etal-2024-investigating} and our approach with various early exiting thresholds, examining generation metrics across a wide range of speedup trade-offs. For Mixture of Depths~\cite{raposo2024mixture}, we conducted experiments with varying layer capacities to assess generation performance as a function of layer capacity and in turn speedup. In the case of Skip Decoding ~\cite{delcorro2023skipdecode}, we adapted exit points according to sequence length to achieve similar trade-offs between generation performance and speedup. 
 Benefits of better alignment (see ~\cref{fig:koala_alignment}) and decision boundaries (see ~\cref{fig:roc_arla}) translate into significantly better generation metrics vs speedup trade-offs in case of M2R2 relative to other approaches as observed in ~\cref{fig:ea_comparison}. This trend is also consistent in instruction tuning setup where our approach achieves better generation metrics to speedup trade-off across all instruction test sets. 
 Interestingly we observe that approaches that are shown to perform well during pre-training such as Mixture of Depths ~\cite{raposo2024mixture} and Skip decoding ~\cite{delcorro2023skipdecode} tend to perform poorly during supervised fine-tuning and instruction-tuning setups. In ~\cref{mod_discountinuity} we provide some empirical reasoning that may lead to this suboptimal behavior. We also measured trainable parameter overhead associated with routers, projection layers and ARLA mechanism described in ~\cref{method_arla} of all approaches to evaluate inference feasibility on resource constrained devices as shown in ~\cref{fig:parameter_overhead}. Our approach uses significantly less parameters since accelerator adapters work fairly well with lower ranks (see ~\cref{fig:adapter_rank_ablation}), while ARLA latent dimesion is substantially low (see ~\cref{method_arla}). Parameter overhead of LITE ~\cite{varshney-etal-2024-investigating} and Skip Decode ~\cite{delcorro2023skipdecode} approaches is dominated by projection layer while that of MoD ~\cite{raposo2024mixture} is dominated by router parameters which include a linear projection and binary classifier at each layer. 

% Furthermore, we note that on compute bound hardware settings where flops are a bottleneck, training change described in \cite{method_training} enables our approach to achieve better flops to speedup trade-off despite consuming more flops per layer demonstrating that benefits of better alignment outweigh flop overhead per layer required accelerate residual streams and obtain better alignment. 

\begin{figure}[ht!]
    \centering
    % First subfigure
    \begin{subfigure}{0.52\textwidth}
        \centering
        \includegraphics[width=\textwidth,height=4cm]{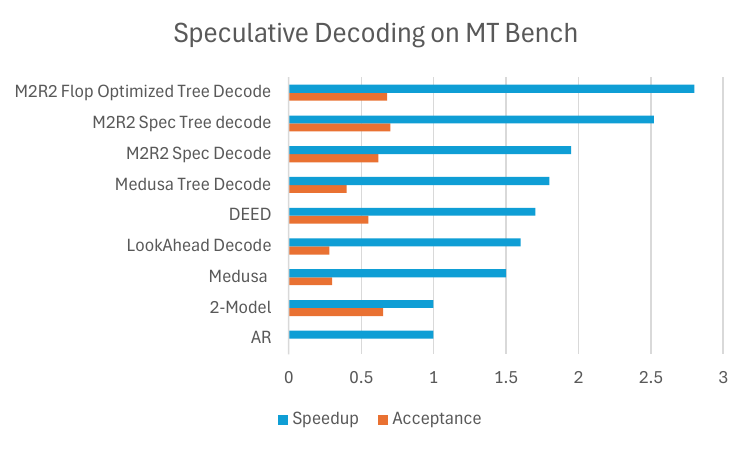}
        \caption{Acceptance rates and speedups of different speculative decoding architectures on Nvidia-A100.}
        \label{fig:sd_speedup}
    \end{subfigure}
    \hfill
    % Second subfigure
    \begin{subfigure}{0.46\textwidth}
        \centering
        \includegraphics[width=\textwidth,height=4cm]{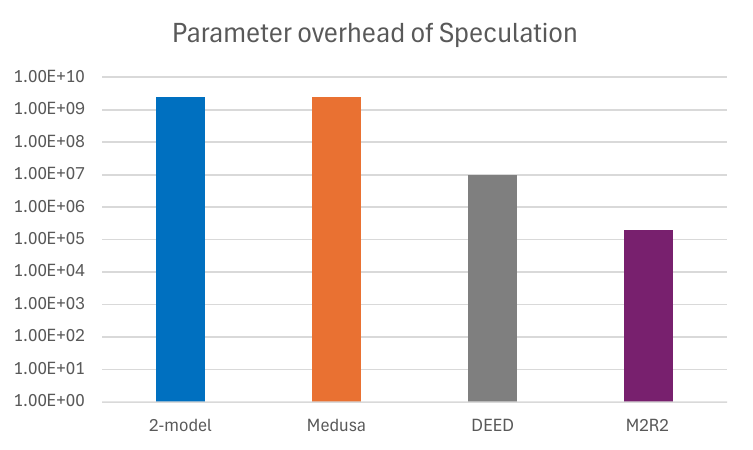}
        \caption{Overhead of trainable parameters of different speculative decoding architectures.}
        \label{fig:sd_overhead}
    \end{subfigure}
    \caption{Effectiveness of M2R2 in Speculative Decoding settings.}
    \label{fig:sd_results}
\end{figure}

\subsection{Speculative Decoding} Improved early alignment significantly boosts acceptance rates when using accelerated residual streams for speculative candidate sampling in speculative decoding settings. As shown in \cref{fig:sd_results}, we compare acceptance rates by sampling tokens from the first $k = 4$ layers of Gemma-7B ~\cite{gemma2024} with our approach, which employs a residual transformation rate $R = N/k$, where $N$ denotes number of layers of base model. Baselines include draft-model-based speculative decoding \cite{leviathan2023fast, chen2023accelerating} that utilizes Gemma-2B \cite{gemma2024} as the draft model, as well as single-model methods such as Medusa \cite{medusa},  DEED \cite{Tang2024} and LookAhead Decoding ~\cite{fu2023lookahead}. For these experiments, we fix the candidate speculation length $\gamma = 3$ and set $k = 4$.

While aligned draft model in 2-model setup achieves high acceptance rates, it does not yield generation speedups on A100 GPUs; the latency of speculative candidate generation outweighs the benefits of parallel acceptance for multiple tokens. In contrast, Medusa performs candidate speculation in a non-autoregressive (NAR) manner, minimizing speculation overhead; however, the acceptance rates are suboptimal due to the lack of dependency between generated speculative tokens \cite{hydra, bhendawade2024speculative}. The method proposed in DEED \cite{Tang2024} generates candidate speculations using only a few layers, making each token dependent on the previous, which incurs low candidate generation costs since only a subset of the model layers is engaged. Our approach addresses lower acceptance rates of DEED ~\cite{Tang2024} by leveraging accelerated residuals, which improve both alignment and generation speedups. Furthermore, as shown in \cref{fig:sd_overhead}, parameter overhead of our approach for speculative draft generation is substantially lower than other baselines  making it an ideal approach for resource-constrained scenarios.

% \begin{figure}[h!]
%     \centering
%     % First subfigure
%     \begin{subfigure}{0.48\textwidth}
%         \centering
%         \includegraphics[width=\textwidth]{sections/figures/moe_hit_rate_koala.png}
%         \caption{Expert speculation hit rate on Koala}
%         \label{fig:subfigure1}
%     \end{subfigure}
%     \hfill
%     % Second subfigure
%     \begin{subfigure}{0.48\textwidth}
%         \centering
%         \includegraphics[width=\textwidth]{sections/figures/moe_inference_time_per_token_koala.png}
%         \caption{MoE inference time per step on Koala}
%         \label{fig:subfigure2}
%     \end{subfigure}
%     \caption{Overall caption for both images}
%     \label{fig:mainfigure}
% \end{figure}

\begin{figure}[h!]
    \centering
    % First subfigure
    \begin{subfigure}{0.48\textwidth}
        \centering
        \includegraphics[width=\textwidth]{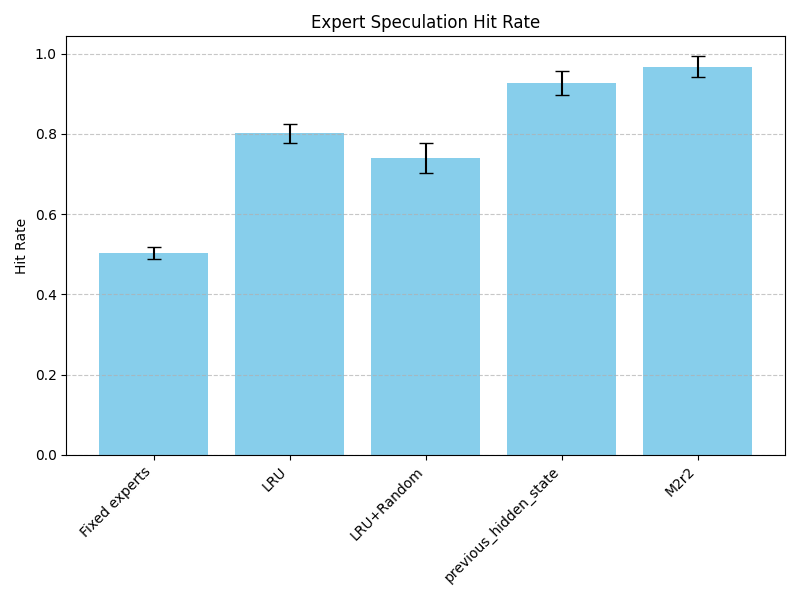}
        \caption{Expert speculation hit rate of different expert speculation strategies on measured on MT Bench.}
        \label{fig:subfigure1}
    \end{subfigure}
    \hfill
    % Second subfigure
    \begin{subfigure}{0.48\textwidth}
        \centering
        \includegraphics[width=\textwidth]{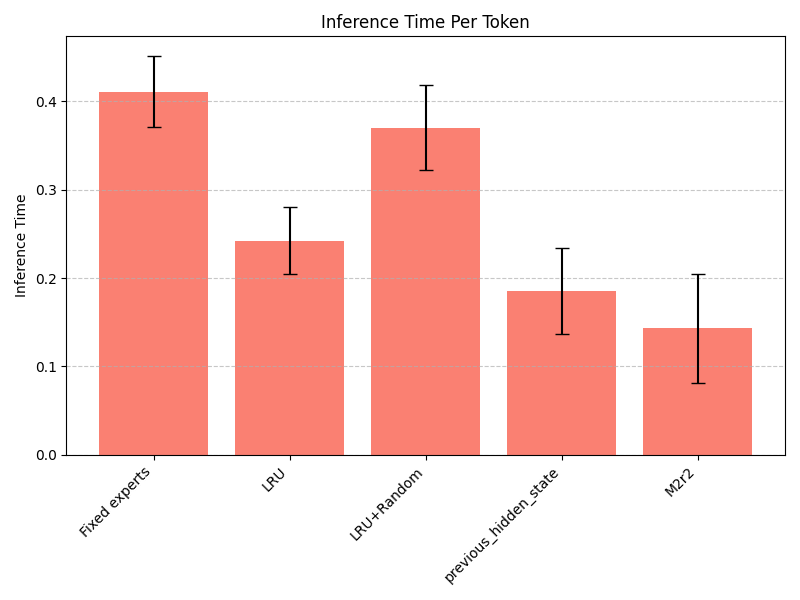}
        \caption{Inference latency per decode step using different expert pre-loading strategies measured on MT Bench.}
        \label{fig:subfigure2}
    \end{subfigure}
    \caption{Effectiveness of M2R2 for speculative expert pre-loading on sparse MoE Transformers }
    \label{fig:MoE_mt_bench}
\end{figure}

\subsection{Ahead-of-Time (AoT) MoE Expert Loading} \label{experiments_aot}

We evaluate efficacy of Ahead-of-Time (AoT) expert loading in resource-constrained environments, where limited high-bandwidth memory (HBM) restricts the number of experts that can be stored in fast-access memory. When an MoE gate selects an expert that is not in HBM, it must be loaded from low-bandwidth memory (LBM), introducing latency. To simulate this scenario on an A100 GPU, we restrict HBM capacity to 8GB, requiring experts beyond this limit to be loaded from LBM.\footnote{In our experimental setup, GPU DRAM is treated as HBM and disk as LBM, though the setup generalizes to architectures where SRAM serves as HBM and DRAM as LBM, a common design in many accelerators ~\cite{jouppi2017tpu, keckler2011gpu, lane2020apple}}.

We compare five strategies to maximize compute efficiency and reduce latency. The first approach fixes experts in HBM without replacement. In the second strategy, we employ a Least Recently Used (LRU) eviction policy, where the least accessed experts are dynamically replaced when new ones are needed. The third method extends the LRU approach by adding speculative caching and randomly pre-loading experts. The fourth strategy uses speculative loading based on residual states from the previous layer. Finally, our proposed approach, described in \cref{method_aot_expert_loading}, leverages accelerated residual states to more effectively speculate and pre-load experts in advance. Using the OLMoE model, which has 64 experts per MLP layer, our 8GB HBM capacity allows only 32 experts to be cached in high-speed memory, while the remaining 32 reside in LBM. 

% \textbf{[Clarification Needed: Confirm if 32 experts per layer correspond to 16GB]}

As shown in \cref{fig:MoE_mt_bench}, both our method and the approach of speculating experts based on residual state of previous layer achieve significantly higher hit rates compared to the fixed and LRU-based strategies. Our method operates on accelerated residuals at a rate of 2X, initiating speculative pre-loading of experts at layers $2i+2$ and $2i+3$ while the GPU kernel is engaged in computing the attention and MLP transformations for layer $i$. Starting pre-loading earlier proves advantageous, as we observe that miss rates tend to increase in the final layers when using LRU caching strategies (see ~\cref{fig:miss_rates}). If speculative pre-loading is initiated only one layer before the current layer, it often results in insufficient loading time, preventing all the necessary experts for layer $i$ from being fully loaded during the computation of layer $i-1$. By pre-loading ahead, our method ensures that most speculated experts are readily available, thereby reducing latency and improving inference efficiency. While we demonstrate the effectiveness of operating at a 2X rate for initiating expert pre-loading, the optimal extent of early pre-loading necessary to maximize inference performance remains an open question. We leave this exploration for future work.

% We also perform ablations on hit rates using a linear mapping to speculate experts early relative to using accelerated residuals in ~\cref{ablations:moe_speculation}. 

\section{Related Work} \label{sec:related}

The inference speed of large language models (LLMs) is often constrained by the sequential nature of auto-regressive decoding, which necessitates a complete forward pass of the network for each token generated. To mitigate the high inference latency associated with LLMs, various strategies have been proposed to reduce their memory footprint. Techniques such as model quantization \cite{frantar2022gptq,yao2022zeroquant,dettmers2023spqr}, knowledge distillation to smaller models \cite{gu2023knowledge,agarwal2023gkd}, and pruning \cite{frantar2023sparsegpt,sun2023simple} have emerged as effective solutions. However, these strategies often neglect the variational complexity inherent in each token, resulting in a reliance on static computation for all tokens. To better address this issue, several early exiting approaches have been developed to facilitate dynamic computation. These methods focus on terminating residual transformations early for simpler tokens, achieving significant speedups in embedding models \cite{xin2020deepspeed,hou2020dynabert,varshney2022model}.
In the context of sequence generation models, techniques like Confident Adaptive Language Modeling (CALM) \cite{schuster2022confident} and Depth-Adaptive Transformers \cite{elbayad2020depthadaptive} have effectively employed early exiting by integrating classifiers into the decoder layers. However, these approaches are constrained by key-value (KV) cache mismatches that arise between the training and inference phases, as KV states are not accessible for tokens that are early-exited. To mitigate these limitations, skip decoding \cite{delcorro2023skipdecode} has been introduced. This method allows for bypassing a progressively increasing number of layers based on the token’s position in the decoded sequence. While this approach effectively circumvents KV mismatches, the pre-defined limitations on the number of bypassed layers can lead to suboptimal generation quality.

Another promising direction involves conditioning residual transformations at each layer through the use of a router. For example, CoLT5 \cite{ainslie2023colt5} employs conditional routing to determine whether a token should follow a heavy or light computational pathway for each feedforward layer in encoder-decoder models. Mixture-of-depths \cite{raposo2024mixture} builds upon this idea by introducing a predictive router at each layer, which enables efficient inference for conditional computation in decoder-only models. Although conditional routing demonstrates potential during pre-training, as illustrated in \cref{sec:experiments}, its effectiveness during supervised fine-tuning and instruction tuning remains limited. This restricts the applicability of this technique across a wider array of publicly available pre-trained models.

Speculative decoding (SD) has also emerged as a potent method for accelerating autoregressive inference. Techniques such as the original SD framework \cite{leviathan2023fast,chen2023accelerating} utilize a smaller draft model to generate token candidates, which are subsequently validated by the target model, achieving speedups of 2-3x. However, this dual-model approach complicates deployment, as it necessitates hosting both models in memory, which can be resource-intensive in constrained environments. Alternatives like Medusa offer single-model solutions but are limited by their inability to account for token dependencies. In contrast, our approach introduces dependencies between speculative tokens, resulting in more coherent and efficient speculation, thereby achieving higher decoding speedups.

The recent proliferation of Mixture-of-Experts (MoE) language models builds on a long-established concept ~\cite{jacobs1991adaptive, jordan1994hierarchical} of training ensembles of specialized models or ``experts,'' and employing a gating function to select the appropriate expert for a given task. Shazeer et al.~\cite{shazeer2017outrageously} further this idea by developing a sparsely gated Mixture-of-Experts language model. Numerous studies have since explored the application of MoE architectures in Transformer-based models for tasks such as machine translation \cite{lepikhin2021gshard}, masked language modeling \cite{fedus2021switch}, and general-purpose LLMs \cite{du2022glam}. Recently, state-of-the-art sparse Mixture-of-Experts models, such as Mixtral-8x7B \cite{jiang2024mixtral} and OLMoE \cite{muennighoff2024olmoe}, have been released, outperforming their open-source dense transformer counterparts across several benchmarks.

% For efficient inference in MoE models, expert caching strategies, such as least-recently-used (LRU) caching, have been proposed to optimize expert utilization. For example, \cite{eliseev2023fast} utilizes residual states from previous layers to inform expert selection; however, the resultant speedups are limited by their reliance on earlier layers for initialization. Expert loading for the current layer can only be activated after the attention computation of the previous layer is complete. Our approach overcomes these constraints by enabling anticipatory expert selection mechanisms through the use of accelerated residuals.

\section{conclusion}
In this paper, we introduced the Mixture of Multi-rate Residuals (M2R2) framework, which dynamically modulates the velocity of residual transformations to optimize early residual alignment, improving inference efficiency in diverse inference setups. Unlike traditional methods that focus on the "distance" tokens traverse within the model layers, M2R2 enhances the rate at which residuals evolve, allowing for faster alignment of intermediate representations. Our empirical results demonstrate that M2R2 outperforms state-of-the-art dynamic computation methods offering better generation metrics to speedup trade-off. Furthermore, it achieves 2.8X speedup in lossless self-speculative decoding setup. In Mixture-of-Experts (MoE) models, with ahead-of-time expert loading, M2R2 reduces decoding latency by overlapping memory transfers with computation and achieves a throughput improvement of up to 2.9X compared to traditional expert loading methods. Overall, M2R2 offers an effective solution for optimizing inference in resource-constrained environments, enhancing both dense transformer and sparse MoE models.

\section*{Acknowledgements}
We would like to thank
Qingqing Cao, Yichen Jiang and Thomas Merth
for their valuable feedback and discussions.

\bibliographystyle{plain}  
\bibliography{m2r2}
\clearpage
\appendix  % Start the appendix

\section{Dynamic Computing}
\subsection{Gradient conflict resolution} \label{sec:grad_conflict}

Traditional early exiting strategies frequently encounter issues related to gradient conflicts ~\cite{predictive_exit_conflicts, gradient_projection_exit}, where multiple exit points induce conflicting gradients during the training phase. This phenomenon leads to optimization instability and challenges in convergence, as gradients computed from divergent branches may not align effectively, and the presence of early exits can perturb the gradient flow, potentially resulting in the incomplete training of lower early exit heads. To illustrate this problem, consider a trainable parameter \( w_j \) situated between gates \( E_j \) and \( E_{j+1} \). For the loss associated with the early exit at gates \( E_{j+1 ... n} \), the parameter update required in \( w_j \) can be expressed as:

\begin{equation}
    \label{eq_grad_propagation_vaniila-ea}
\Delta w_{j} = -\eta \sum_{k=j+1}^{n} \beta_k \frac{\partial L_{E_k}}{\partial w_k}
\end{equation}

where \( \beta_k \) is the backward transformation coefficient for the gradient from gate \( E_k \) to reach parameter \( w_{j} \) and \( \eta \) is the learning rate. Conversely, since accelerated residuals at gate $E_j$ are initialized from slow residuals $H_j$ which are trained with base adapters, when base adapters are frozen, gradient propagation is limited to parallel adapter parameters from gate $E_j$ to gate $E_{j+1}$ thus ensuring every parallel adapter parameter is optimized for specific exit as shown in ~\cref{fig:m2r2_main}. Formally speaking, the update of accelerator adapter parameter $w_j$  within our proposed framework is delineated as:

\begin{equation}
    \label{eq_grad_propagation_ss}
    \Delta w_{j} = 
\begin{cases} 
 -\eta \hat{\beta}_{j+1} \frac{\partial L_{E_{j+1}}}{\partial w_j} & \text{if } E_j < w_j < E_{j+1} \\
0 & \text{otherwise}
\end{cases}
\end{equation}

where \( \hat{\beta}_{j+1} \) is the backward transformation coefficient for the gradient from gate \( E_{j+1} \) to reach parameter \( w_{j} \) of accelerator adapter. This formulation mitigates gradient conflicts arising from gradients associated with top gates, thereby enhancing the stability of the optimization process.\footnote{For simplicity, we focus on cross-entropy loss in this discussion; however, the same reasoning extends to distillation loss as detailed in ~\cref{method_training}.}

\begin{figure}[ht]
    \centering
    \begin{subfigure}{0.48\textwidth}
        \centering
        \includegraphics[width=\textwidth]{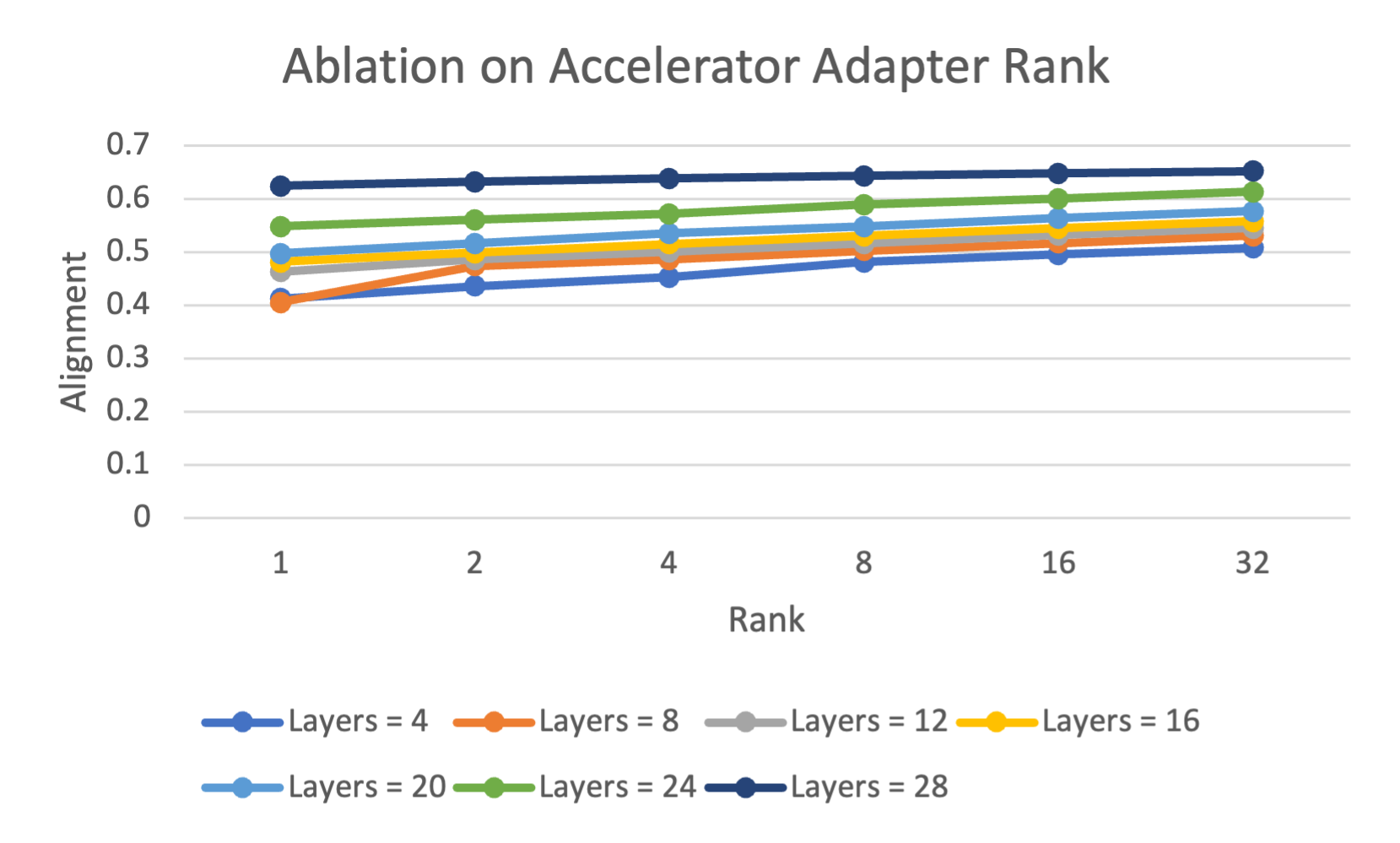}
        \caption{Alignment of early-exited tokens with those from the final layer improves as adapter rank increases, but tends to plateau beyond a rank of 8. }
        \label{fig:adapter_rank_ablation}
    \end{subfigure}%
    \hfill
    \begin{subfigure}{0.48\textwidth}
        \centering
        \includegraphics[width=\textwidth]{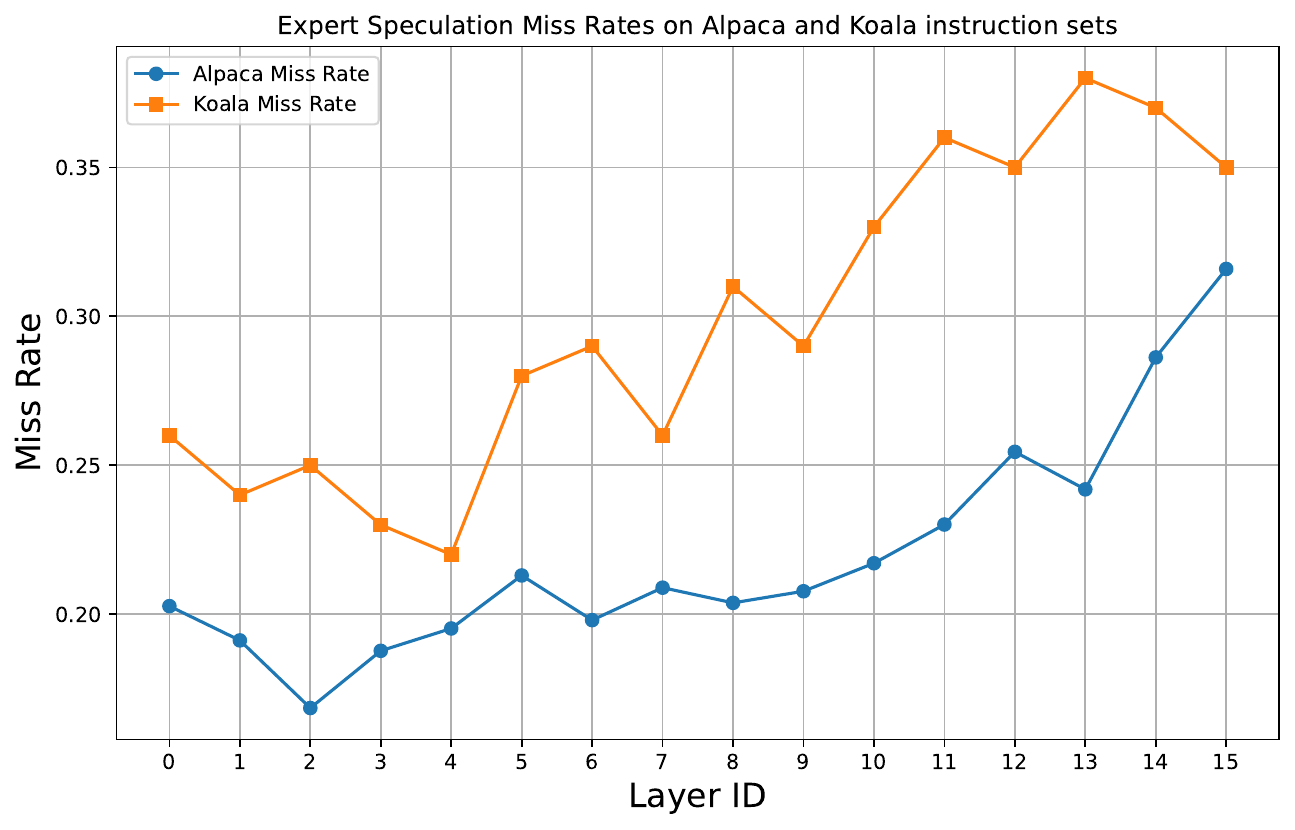}
        \caption{Expert speculation miss rates increase in later layers with LRU policy, while initial layers exhibit lower miss rates. Thus, accurate speculative pre-loading benefits later layers more. We leverage accelerated residuals to speculate and pre-load experts for these layers during the computation of earlier layers.   }
        \label{fig:miss_rates}
    \end{subfigure}
    \caption{(a) Adapter Rank Ablation on Dialog Summarization (b) Expert speculation miss Rates}
    \label{fig:overall_fig}
\end{figure}

 \subsection{Discontinuity in Mixture of Depths and Skip Decoding} \label{mod_discountinuity}

To get a deeper understanding of discontinuity leading to suboptimal performance of architectures like MoD and Skip decoding during instruction tuning and fine-tuning phases, we pass a diverse set of prompts through the models and observe residual stream transformation. Residual streams in pre-trained dense transformers tend to undergo significant changes in both direction and magnitude in first few layers than later layers as depicted in \cref{fig:residual_change}. Since Mixture of Depth (MoD) relies on skipping the residual transformation for some of the tokens determined by router, skipping early transformations makes it harder to obtain final residual state closer to that obtained from full dense transformers even when MoD parameters are explicitly trained to alleviate this discontinuity. Skip decoding on the other hand approximates skipping residual transformation of first few layers with a linear projection while  ignoring non-linearities and context, leading to sub-otpimal performance as well.

\section{MoE Speculation Continued}

In this section, we detail the expert transfer process between High Bandwidth Memory (HBM) and Low Bandwidth Memory (LBM) on the A100 GPU. We employ CUDA’s multi-stream functionality ~\cite{cuda_programming_guide} to establish distinct compute and memory-loading streams, both of which operate concurrently during each forward pass. The load stream is scheduled ahead of the compute stream to ensure efficient memory management: while the compute stream processes layer \(i\), the load stream transfers the least recently used experts of layer \(2i+2\) and \(2i+3\) to LBM and loads speculated experts into HBM. This approach leverages the accelerated residual at layer \(i\), which exhibits strong similarity to the slow residuals at layers \(2i+2\) and \(2i+3\) (see ~\cref{fig:m2r2_residual_sim}), enabling effective expert speculation as shown in ~\cref{fig:moe_expert_aot_loading}. Before executing the MLP experts, we verify whether all required experts are available on HBM; if not, the load stream initiates prioritized, on-demand loading for the experts necessary for MLP computation at layer \(i\). Coordination between the load and compute streams is managed using CUDA primitives.

\begin{figure}
    \centering
    \includegraphics[width=1.0\linewidth]{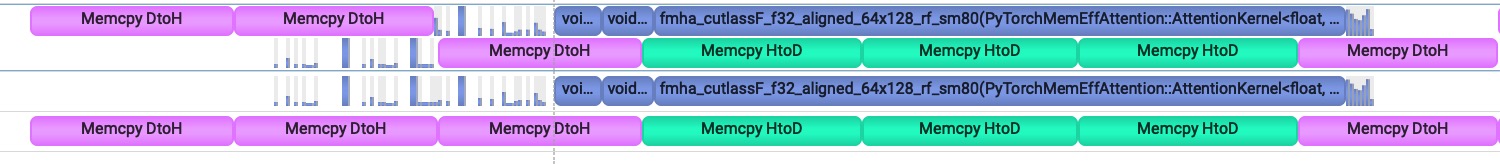}
    \caption{A100 GPU trace demonstrating overlap of computation and expert transfer between LBM and HBM. }
    \label{fig:moe_aot_cuda_trace}
\end{figure}

\begin{figure}[h]
    \centering
    \begin{subfigure}{0.48\textwidth}
        \centering
        \includegraphics[width=\linewidth]{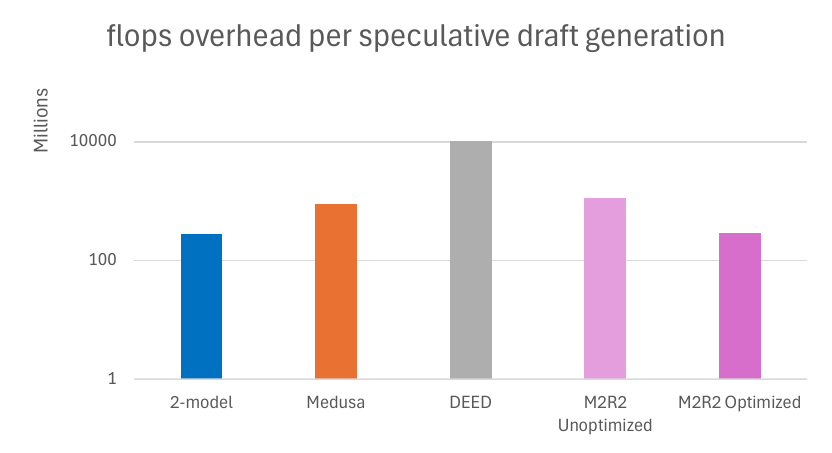}
        \caption{FLOPs overhead of generating speculative draft for different approaches on Gemma-7B. The optimized M2R2 approach incorporates FLOPs optimization techniques described in \cref{sec:flops_optimization}}
        \label{fig:flops_optmization}
    \end{subfigure}
    \hfill
    \begin{subfigure}{0.48\textwidth}
        \centering
        \includegraphics[width=\linewidth]{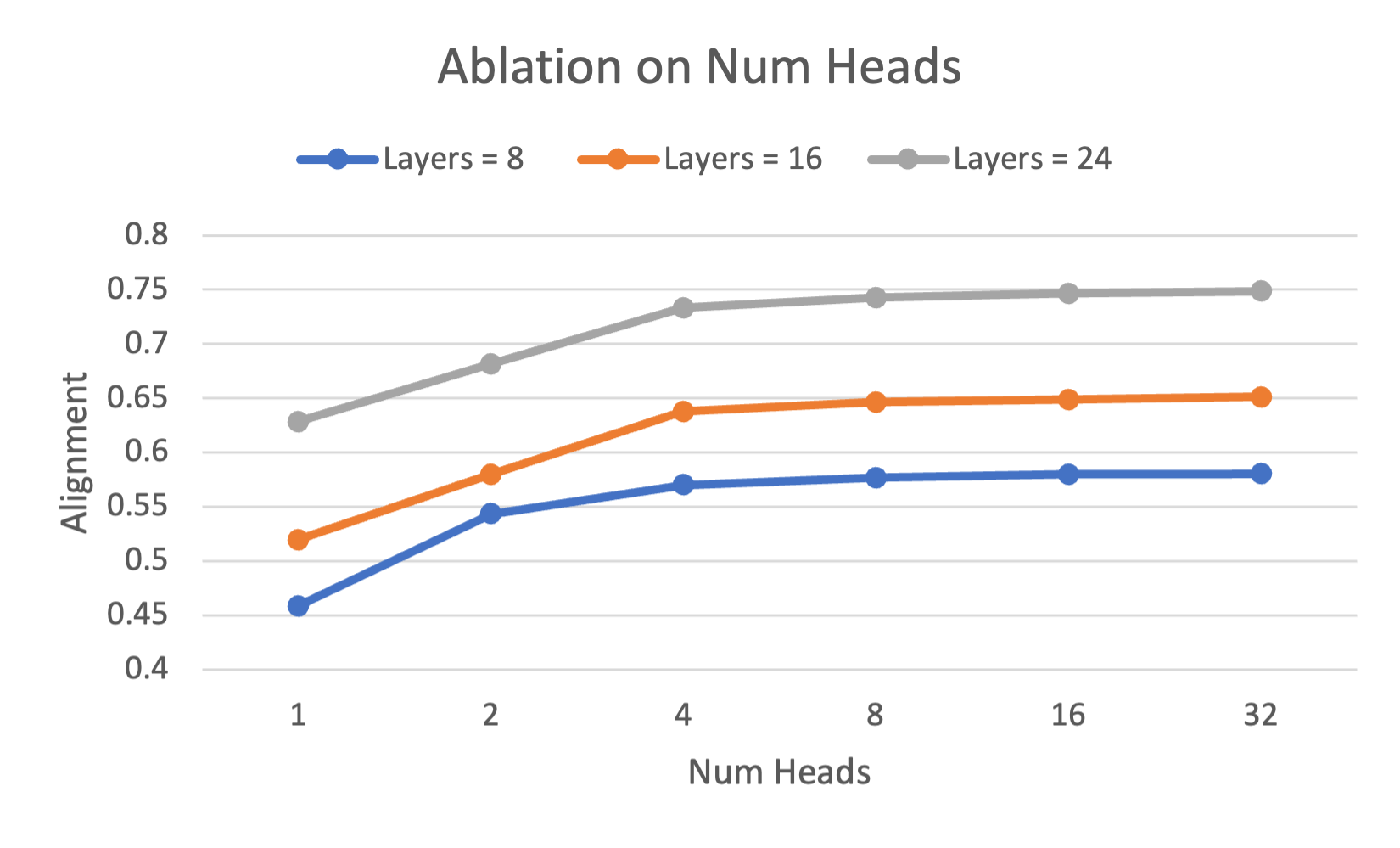}
        \caption{Ablation study on attention heads and M2R2 alignment benefits. Using 4 to 8 heads in the accelerated residual stream reduces FLOPs with minimal alignment degradation.}
        \label{fig:m2r2_num_heads_ablation}
    \end{subfigure}
    \caption{FLOPs overhead of M2R2 and optimization based on Attention head pruning.}
\end{figure}

% \begin{figure}[htbp]
%     \centering
%     \includegraphics[width=0.5\linewidth]{sections/figures/adapter_rank_ablation.png} 
%     \caption{Adapter Rank Ablation on Dialog Summarization}
%     \label{fig:adapter_rank_ablation}
% \end{figure}

% \section{FLOP Efficient Mode} \label{appendix:flop_efficient_mode}
% Transformer-based decoding is generally memory-bound on most mainstream accelerators ~\cite{liang2020transformer}; however, there exist specific deployment scenarios where reducing floating-point operations (FLOPs) is essential. For example, in on-device settings, power consumption is often directly proportional to the number of FLOPs per decoding step, making FLOP reduction a critical factor in minimizing energy consumption. 
% Since M2R2 uses a parallel faster stream during each forward pass, FLOPs especially during attention computation increase substantially, resulting in overall more flops than SOTA approaches as shown in ~\cref{fig:m2r2_flops_overhead}. To operate in Flop efficient mode, we propose a slimmer accelerated residual stream that uses a fraction of attention heads as that of slower residual stream. Specifically, we use first $\hat{n}_h$ heads of query, key and value projections of base model to process a slicker accelerated stream. ~\cref{fig:m2r2_num_heads_ablation} indicates effect of using a slicker stream on alignment. As depicted, using $\hat{n}_h = 8$ offers a good trade-off between alignment and FLOPs overhead. 

% So we used $\hat{n}_h = 4$ for all experiments described in ~\cref{sec:experiments}

\textbf{Accelerator Adapter Rank Ablation} \label{m2r2_adapter_rank}
To minimize parameter overhead from accelerator adapters, we conduct an ablation study on adapter rank to identify the optimal rank that achieves strong alignment without substantially increasing parameter load. As illustrated in~\cref{fig:adapter_rank_ablation}, a rank of 8 offers an effective trade-off, with alignment performance showing a steep improvement up to rank 8, beyond which the benefit curve begins to plateau.

\newpage
\textbf{Prompt for Evaluation of Dynamic Compute Responses} \label{m2r2_prompt}

To assess the responses generated by our approach alongside baseline models, we utilize the following prompt for GPT-4 oracle. Note that the baseline and target responses are randomly assigned to either Assistant 1 or Assistant 2 in the template below.

\begin{lstlisting}[basicstyle=\ttfamily, breaklines=true]
Human: You are a helpful and precise assistant for evaluating the quality of an answer.
[Question]
{question}
[The Start of Assistant 1's Answer]
{answer_1}
[The End of Assistant 1's Answer]
[The Start of Assistant 2's Answer]
{answer_2}
[The End of Assistant 2's Answer]

We request your feedback on the performance of both AI assistants in response to the user question above. Please rate their responses based on helpfulness, relevance, accuracy, and level of detail.

Assign each assistant an overall score on a scale of 1 to 10, where a higher score reflects better performance.

Please provide a single line output with only two values, representing the scores for Assistant 1 and Assistant 2, respectively, separated by a space.

Assistant:
\end{lstlisting}

\end{document}